\documentclass{article}
\PassOptionsToPackage{hyphens}{url}

\usepackage{microtype}
\usepackage{graphicx}
\usepackage{subcaption}
\usepackage{booktabs}
\usepackage{hyperref}
\usepackage[normalem]{ulem}
\usepackage{xcolor}

\usepackage[accepted]{icml2026}

\usepackage{amsmath}
\usepackage{amssymb}
\usepackage{mathtools}
\usepackage{multirow}
\usepackage[T1]{fontenc}
\usepackage{listings}
\usepackage{tikz}
\usetikzlibrary{arrows.meta,positioning,shapes.geometric,fit,backgrounds}

\usepackage[capitalize,noabbrev]{cleveref}

\icmltitlerunning{When Does Critique Improve AI-Assisted Theoretical Physics? SCALAR: Structured Critic--Actor Loop for Agentic Reasoning}

\begin{document}

\twocolumn[
    \noindent
     CCTP-2026-7 \hfill CERN-TH-2026-097\\
\noindent
    ITCP-2026-7 \hfill QMUL-PH-26-15
    \vspace{0.2em}

  \icmltitle{When Does Critique Improve AI-Assisted Theoretical Physics?\\
    SCALAR: Structured Critic--Actor Loop for Agentic Reasoning}

  \icmlsetsymbol{equal}{*}

  \begin{icmlauthorlist}
    \icmlauthor{Vasilis Niarchos}{crete}
    \icmlauthor{Constantinos Papageorgakis}{qmul}
    \icmlauthor{Alexander G. Stapleton}{qmul}
    \icmlauthor{Sokratis Trifinopoulos}{cern,uzh}
  \end{icmlauthorlist}

  \icmlaffiliation{crete}{Department of Physics, CCTP and ITCP, University of Crete, 71303, Greece}
  \icmlaffiliation{qmul}{Centre for Theoretical Physics and Astronomy, School of Physical and Chemical Sciences, Queen Mary University of London, London E1 4NS, United Kingdom}
  \icmlaffiliation{cern}{Theoretical Physics Department, CERN, Geneva, Switzerland}
  \icmlaffiliation{uzh}{Physik-Institut, Universit\"at Z\"urich, 8057 Z\"urich, Switzerland}
  
  \icmlcorrespondingauthor{Sokratis Trifinopoulos}{sokratis.trifinopoulos@cern.ch}

  \icmlkeywords{LLM agents, physics reasoning, multi-agent systems, prompting strategies}

  \vskip 0.3in
]

\printAffiliationsAndNotice{}

% ============================================================
\begin{abstract}                          
As large language models (LLMs) show increasing promise on research-level physics reasoning tasks and agentic AI becomes more common, a practical question emerges:
\emph{How does the interaction between researchers and agents affect the results?}
We study this using SCALAR (Structured Critic--Actor Loop for Agentic Reasoning), an Actor--Critic--Judge pipeline applied to quantum field theory and string theory problems. The Actor proposes solutions, the Critic provides iterative feedback, and an independent Judge evaluates the transcript against reference solutions. We vary the Actor persona, the Critic feedback strategy, and the Actor model family and scale. Multi-turn dialogue improves over single-shot attempts throughout, but both the mechanism of improvement and the value of different prompting choices depend strongly on the Actor--Critic pairing. Increasing the scale within one model family (\emph{e.g.} from the 8B-parameter DeepSeek-R1 variant to DeepSeek-R1 70B) improves some easier-problem behavior, but does not remove the hardest bottleneck we observe. Critic feedback strategy matters most clearly in the asymmetric Actor--Critic setting (\emph{e.g.}, a lightweight  Haiku Actor guided by a stronger Sonnet Critic), where constructive feedback improves mean-score outcomes. In same-family Actor--Critic settings, strategy effects are weaker: lenient feedback is sometimes favored, while strict and adversarial feedback are not beneficial. Taken together, SCALAR provides a controlled testbed for evaluating which interaction structures help or hinder AI-assisted reasoning on known-answer scientific problems.
  \end{abstract} 
 
% ============================================================
\section{Introduction}
\label{sec:intro}

Large language models (LLMs) and LLM-based agents are a new type of interlocutor in the dialogue that drives the scientific process. They can reason, make decisions (rather than merely perform algorithmic computations), and even exhibit greater adaptability to iterative prompts than one-shot queries. This behavior is closer to that of a human collaborator than that of any previous computational tool. Early evidence of new contributions to theoretical physics~\citep{guevara2026singleminus,Schwartz:2026ekw,Shih:2026lmy,Shih:2026jfe,lu2025languageagentsphysics}, mathematical discovery~\citep{romera2024mathematical,Novikov:2025alphaevolve}, and agentic scientific workflows in high-energy physics~\citep{Plehn:2026gxv,Agrawal:2026lvg} is very encouraging. However, how physicists should
structure this collaboration is an open question: in general, multi-turn interactions exhibit
sticky error states and capability degradations 
~\citep{liang2024mathchatbenchmarkingmathematicalreasoning,laban2025lost,zhang2025turnbench}, although structured
multi-agent pooling can reduce hallucinations~\citep{till2025multiconsistency}. 

We probe
these dynamics with SCALAR, a deliberately pedagogical \emph{Actor--Critic--Judge} pipeline: one LLM agent (comparable to a student) plays the \emph{Actor} attempting a graduate-level quantum-field-theory (QFT) or string theory problem. The \emph{Critic} LLM (analogous to a teaching assistant) then delivers formative feedback mid-task, before the final agent playing the \emph{Judge} (teacher) sets the standard against which the work is
ultimately evaluated; for prior discussions of LLMs as Judges see e.g. \cite{NEURIPS2023_91f18a12}. Such a pedagogical interpretation completes a scaffolding in the sense of educational psychology~\citep{wood1976scaffolding,vygotsky1978}; interaction styles are also heavily studied in human dialogs~\citep{chi2001tutoring,shute2008formative,hattie2007feedback}. We therefore use the language of human--AI collaboration in a stylized, HCI-inspired sense: the SCALAR workflow automates the Critic and Judge roles to isolate interaction variables, rather than measuring live human-subject behavior~\citep{amershi2019guidelines,shneiderman2020human}. Multi-agent approaches to LLM reasoning have shown promise --- from debate
frameworks~\citep{du2024improving,estornell2024accdebate} to specialized refinement
agents for physics~\citep{jaiswal2024mora} and interpretable AI-scientist
collaboration~\citep{xu2025multiagentphysicist}. Pre-prompting strategies have also been shown to make measurable differences to the perceived quality of the LLM's output~\citep{gupta2024persona}, and recent work on instructional distraction shows that models can be sensitive to how task instructions are embedded in surrounding text~\citep{hwang2025instructionaldistractions}. To our knowledge, however, no prior work has systematically studied
which \emph{interaction strategies} between human and AI lead to the best outcomes in the field of theoretical physics.

Our motivation
is threefold and separate from the controlled design choices below, so that exploratory observations are not confused with general prescriptions. First, physicists are already consulting LLMs for
their day-to-day calculations, meaning one should evaluate not only the first answer, but the
interaction as a whole. Moreover, one should ask how reliably agent responses
converge, how they respond to challenge, and where they fail. This is
prerequisite to calibrating that usage in a regime not assessed by single-turn
benchmarks~\citep{chung2025tpbench, gao2025testtimescaling,zhang2025physreason}.
Second, theoretical research is moving towards workflows in which the
physicist supervises a collection of AI agents, rather than
interacting with one model at a time. In our automated benchmark this
supervisory role is stylized as an external Judge, while moment-to-moment
Critic feedback is delegated to an AI teaching assistant. In open-ended use,
the physicist may instead occupy part of the Critic role directly by probing
claims, supplying consistency checks, and deciding when the exchange has met
the required standard. Understanding which scaffolding styles help which AI
actors reach correct solutions is therefore a prerequisite for making either
supervisory mode productive.
Third, SCALAR gives us a controlled testbed in which to hypothesis-test widely repeated prompt-engineering claims --- \emph{e.g.},\ the report that
``assigning the model a persona'' can swing performance by tens of
percentage points~\citep{gupta2024persona} --- on current-generation models
and on reasoning-heavy scientific tasks.

To study these questions we introduce independent
axes of variation for each party. The Actor 
is varied through an \emph{Actor persona}, defined as the combination
of an \emph{expertise level} (novice, expert, or unspecified default)
and a \emph{reasoning style}
(meticulous, physical, skeptical, or left unspecified). This is the
kind of Actor pre-prompting a physicist might use when asking a model
to approach a problem in a particular way. The Critic is
varied across a range of \emph{Critic feedback strategies} (from lenient
and pedagogical to strict and adversarial, plus an unspecified
default), capturing how the assistant intervenes. During the dialogue, the Judge enters only as a
reference-backed evaluator of correctness, silent with respect to the
Actor--Critic exchange. Stored transcripts can then be re-scored by
additional Judges, letting us separate interaction effects from
Judge-specific scoring effects. This defines the role of the Judge as an authority that sets the standard
of the exchange without actively participating in it.
Each persona--strategy configuration is sampled several times on graduate-level QFT and string theory problems,
allowing us to estimate trends across repeated dialogues rather than relying on single runs.
In addition to endpoint scores and convergence rates, we use per-turn
\emph{score-update curves} as a compact diagnostic of when Critic feedback
continues to move the Actor and when the dialogue's effect on the score has plateaued.

While SCALAR is set up in a theoretical physics context, it can be
straightforwardly extended to other domains. The lessons we are
attempting to extract about LLM pre-prompting and interaction can
eventually inform the optimization of multi-agent setups, where the
agent persona and skill set play an important role. More generally, we
view our analysis as a step towards more efficient AI-assisted
open-ended research; the interaction patterns identified here provide a
vocabulary for that future work.

\begin{figure}[t]
  \centering
  \scalebox{0.6}{
  % \documentclass[tikz,border=8pt]{standalone}
% \usepackage[T1]{fontenc}
% \usepackage{lmodern}
% \usepackage{tikz}
% \usetikzlibrary{arrows.meta,positioning,shapes.geometric,fit,backgrounds}

% \begin{document}
\begin{tikzpicture}[
  node distance=10mm and 14mm,
  >=Latex,
  line/.style={-Latex,thick},
  block/.style={
    rectangle,
    rounded corners=2mm,
    draw,
    thick,
    minimum width=34mm,
    minimum height=10mm,
    text width=34mm,
    align=center
  },
  io/.style={
    trapezium,
    trapezium left angle=72,
    trapezium right angle=108,
    draw,
    thick,
    minimum width=36mm,
    minimum height=10mm,
    text width=36mm,
    align=center
  },
  save/.style={
    trapezium,
    trapezium left angle=72,
    trapezium right angle=108,
    draw,
    thick,
    minimum width=36mm,
    minimum height=10mm,
    text width=36mm,
    align=center
  },
  decision/.style={
    diamond,
    aspect=2.0,
    draw,
    thick,
    minimum width=22mm,
    minimum height=10mm,
    text width=14mm,
    align=center,
    inner xsep=1.5mm,
    inner ysep=1mm
  },
  term/.style={
    ellipse,
    draw,
    thick,
    minimum width=30mm,
    minimum height=10mm,
    align=center
  },
  actor/.style={fill=blue!10},
  critic/.style={fill=orange!14},
  judge/.style={fill=green!14},
  note/.style={font=\scriptsize,align=center},
]

% Core nodes
\node[io] (problem) {Problem and \\Persona/Strategy Setup};
\node[block,actor,below=of problem] (actor) {Actor\\ \textit{Form solution}};
\node[coordinate,above=7mm of actor] (actor_in) {};
\node[block,critic,below=of actor] (critic) {Critic\\\textit{Feedback + \\Error Checks}};
\node[block,judge,below=of critic] (judge) {Judge\\ \textit{Score + Verdict}};
\node[decision,below=of judge] (pass) {Pass?};
\node[decision,right=22mm of pass] (earlystop) {Early\\stop?};
\node[save,below=of pass] (save) {Save\\Run};
\node[term,below=of save] (end) {End};

% Flow edges
\draw[line] (problem) -- (actor);
\draw[line] (problem) -- (actor_in);
\draw[line] (actor_in) -- (actor);
\draw[line] (actor) -- (critic);
\draw[line] (critic) -- (judge);
\draw[line] (judge) -- (pass);
\draw[line] (pass) -- node[note,left,pos=0.55] {\large{Yes}} (save);
\draw[line] (critic.west) -| ++(-18mm,0) |- node[note,near start,left] {\large{Feedback loop}} (actor_in);
\draw[line] (pass.east) -- node[note,above] {\large{No}} (earlystop.west);
\draw[line] (earlystop.south) |- node[note,pos=0.25,right] {\large{Yes}} (save.east);
\draw[line] (earlystop.north) |- node[note,near start,right] {\large{No: iterate}} (actor_in);
\draw[line] (save) -- (end);

% % Background grouping box
% \begin{scope}[on background layer]
%   \node[
%     draw=black!45,
%     rounded corners=2mm,
%     dashed,
%     inner sep=6mm,
%     fit=(actor)(critic)(judge)(pass)(earlystop)(save)(end),
%     % label={[font=\footnotesize]above:Actor-Critic-Judge Loop}
%   ] {};
% \end{scope}

\end{tikzpicture}
% \end{document}
  }
  \caption{The SCALAR Actor--Critic--Judge pipeline. The Actor and the Critic engage in iterative dialogue, while an independent evaluator
    (Judge) scores the Actor's current solution against a ground truth. The Actor's output is
    shaped by the Actor persona; the Critic's feedback is shaped by the
    Critic feedback strategy. 
    }
  \label{fig:pipeline}
\end{figure}
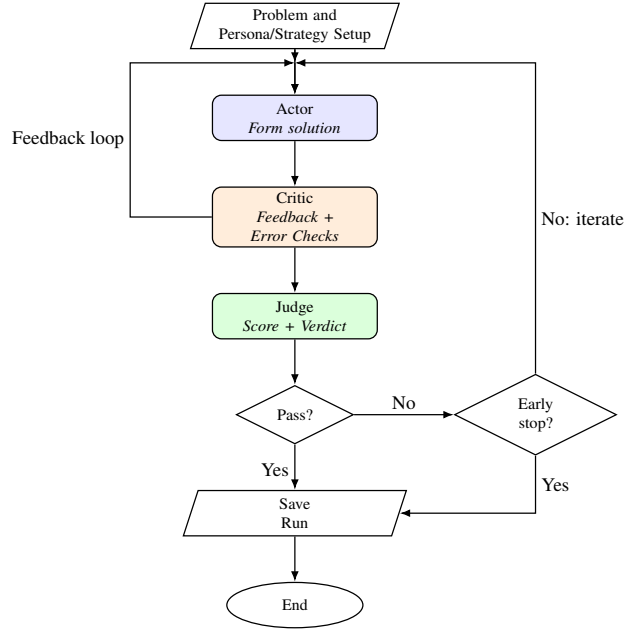

% ============================================================
\section{Methods}
\label{sec:methods}

% ============================================================

\subsection{Roles, Feedback Strategies and Pipeline}
\label{sec:roles}

We begin the discussion of our methods by describing the SCALAR Actor personas and Critic feedback strategies, both of which are implemented through pre-prompting. The design space is intentionally small rather than exhaustive: common persona-prompting choices and tutoring/scaffolding feedback modes are chosen to permit a full factorial sweep.
The Actor persona is factored into two orthogonal dimensions. The first sets the expertise level:
\emph{expert} (``you are an expert in theoretical physics''),
\emph{novice} (``you are a student learning QFT''),
or \emph{default} (no expertise instruction).
The second covers different reasoning styles, which shape the approach the Actor takes in the
calculations:
\emph{meticulous} (emphasizing careful algebra and cross-checks),
\emph{physical} (prioritizing physical intuition and limiting cases),
\emph{skeptical} (questioning assumptions at each step),
or \emph{default} with no style instruction.
The full set of combinations yields $3 \times 4 = 12$ Actor personas.

The Critic's feedback strategy controls the tone of the feedback provided to the Actor:
\emph{adversarial} (aggressively challenging claims),
\emph{strict} (precise error flagging),
\emph{pedagogical} (Socratic questioning),
\emph{lenient} (gentle suggestions accepting partial progress),
and \emph{default} with no stylistic emphasis~\citep{baron1988criticism,butler1987evaluation,shute2008formative,hattie2007feedback}.
The full prompt texts are given in \cref{app:prompts}. The LLM assigned to each role
  is an additional degree of freedom and our model choices are described in \cref{sec:models}.

With the pre-prompting fixed, SCALAR proceeds as follows (\cref{fig:pipeline}).
Given a problem statement and persona--strategy configuration, the Actor produces an initial solution attempt.
The Critic, who has access to a reference solution but is instructed not to disclose it, then reviews this attempt, flags errors, and delivers structured feedback.
The Judge scores the Actor's work against the reference solution and issues a pass/fail verdict.
If the Actor passes, or if an early-stopping criterion is met (iteration limit or score stagnation), the run is saved and terminated.
Without this, the Critic's feedback is passed back to the Actor for a further attempt and the loop repeats until the stopping criteria are met.
For analysis, the recorded state after each Actor turn consists of the
fixed experimental settings and the dialogue record available at that
point. Thus the recorded dialogue state is the natural Markov state of
the generated dialogue: conditional on that state and the fixed
configuration, the next role call is generated without invoking any
additional recorded dialogue history. In \cref{sec:results}, we analyze
the Judge score as a scalar projection of this evolving Markov state.

% ============================================================
\subsection{Evaluation and Metrics}
\label{sec:eval}

At each iteration~$t$ the Judge scores the Actor's current solution in six
dimensions totaling $100$ points: correctness ($50$), mathematical rigor ($10$),
logical flow ($10$), justification quality ($10$), completeness ($10$), and
physical consistency ($10$). To ensure the validity of this grading scheme, we manually reviewed outliers and utilized independent re-scoring as calibration checks. Let $s_t \in [0,100]$ denote the total score at
turn~$t$. Here lowercase $t$ indexes turns, while uppercase $T_i$ denotes
the number of scored Actor states in run~$i$. A run with $T$ scored Actor states produces a sequence
$s_0, s_1, \ldots, s_{T-1}$. The Judge operates outside the Actor--Critic loop: its scores do not feed back into the dialogue, so the same transcripts can be re-scored
by different Judge LLMs to separate dialogue-level effects from judge-specific ones.
Below, a subscript~$i$ denotes a run, and angle brackets denote arithmetic
averages over the indicated set of runs.

We report three evaluation metrics: two score-based quantities measured in points out of
$100$, and one rate that we report as a percentage of runs:
\begin{itemize}
\item \textbf{Mean per-turn score:}
  $\bar{s}_i \;=\; T_i^{-1}\sum_{t=0}^{T_i-1} s_{i,t} \;\in\; [0, 100]$.
  This is the average per-turn score across the whole dialogue. When we quote
  a group mean $\bar s$, we mean the arithmetic mean of these run-level
  quantities. When we instead quote a \emph{final score}, we say so explicitly
  and mean $s_{i,T_i-1}$ averaged over runs.

\item \textbf{Gain:} $g_i = s_{i,T_i-1} - s_{i,0} \in [-100, +100]$. This is the endpoint improvement in
  score points out of~$100$. $g_i>0$ means the dialogue made the solution better.

 \item \textbf{Convergence rate}: Let $r_i \in \{0,1\}$ denote whether the
  run converged. A run is counted as converged when at least one iteration
  of the dialogue produces an Actor solution satisfying all three criteria:
  correctness~$\geq 40$
  (\emph{i.e.},\ $\geq\!80\%$ of the $50$-point correctness rubric), total Actor
  score~$\geq 80$, and final-answer equivalence with the reference, under
  the scoring Judge. For runs whose original loop was driven by the same
  Judge that scores them, the passing iteration is also the terminal
  iteration; for re-scored transcripts the loop length is fixed by the
  original Judge and the re-scoring Judge can mark a non-terminal iteration
  as passing. For any group of runs~$G$, the convergence rate is
  $R_G=\langle r_i\rangle_{i\in G}$ and is reported as a percentage.
  See~\cref{app:metrics} for the formal rule.
\end{itemize}

For cross-problem Critic feedback strategy comparisons we also use
\emph{problem-normalized contrasts}. Let $m(i)$, $p(i)$, and $c(i)$ denote run~$i$'s Actor model
setting (Haiku, DS8B, or DS70B), problem, and Critic feedback strategy. For Actor
model setting~$m$ and Critic feedback strategy~$c$,
define
\begin{align*}
D_{\bar s}(m,c)
&=
\left\langle
\bar s_i-\langle \bar s\rangle_{m,p(i)}
\right\rangle_{m(i)=m,\,c(i)=c},\\
D_R(m,c)
&=
100\left\langle
r_i-\langle r\rangle_{m,p(i)}
\right\rangle_{m(i)=m,\,c(i)=c}.
\end{align*}
Thus $D_{\bar s}$ is measured in score points and $D_R$ in percentage
points. These quantities are descriptive contrasts, not new raw scores: they
ask whether a Critic feedback strategy sits above or below the local Actor--problem baseline.
Here ``local'' means the baseline for the same Actor model setting and the
same problem, after pooling over personas and Critic feedback strategies; the
contrast therefore removes the much larger differences in baseline problem
difficulty before comparing Critic feedback strategies.

\subsection{Problems and Models}
\label{sec:models}

We test SCALAR on three graduate-level QFT and string theory problems drawn from standard textbooks.
\emph{Peskin~\& Schroeder~2.3}~\citep{peskin1995introduction} requires computing the
Feynman propagator at spacelike separation, yielding the modified Bessel function
result $\frac{m}{4\pi^2 r}K_1(mr)$.
\emph{Peskin~\& Schroeder~4.2}~\citep{peskin1995introduction} asks for the lowest-order
lifetime of a heavy scalar particle decaying into two lighter scalars.
\emph{Polchinski 2.7}~\citep{polchinski1998string} requires deriving Operator Product Expansion (OPE) coefficients in the free boson CFT, a core calculation in conformal field theory. Each of these questions was specifically chosen to examine different facets of physics reasoning. For example, Peskin~\& Schroeder 2.3 and Polchinski 2.7 are conceptually straightforward, longer calculations, requiring care with algebraic manipulations, but less physical reasoning and intuition. Conversely, Peskin~\& Schroeder 4.2 is a technically shorter question designed to test conceptual knowledge of decay rates in QFT. For this work the problem set is intentionally chosen to be small, allowing us to deploy many strategies with modest compute. The tradeoff is that the results should be read as a controlled case study of interaction structure between agents rather than as a broad benchmark of theoretical-physics reasoning. Since these are published textbook exercises, we also cannot rule out pretraining exposure to problem statements or solution-manual material. Our claims therefore concern interaction dynamics under possible memorization/contamination, not a clean test of de novo physics knowledge.

We study three Actor model settings and introduce the abbreviations used below. The first uses DeepSeek-R1~70B~\citep{deepseek2025r1} (70B parameters; hereafter DS70B) as both the Actor and Critic, with QwQ-32B~\citep{qwq2024} (32B parameters; hereafter QWQ) as the primary Judge; each of the $12 \times 5 = 60$ persona--strategy configurations is run with 15~repeats (5~per problem) at temperature $T{=}0.7$ for up to 4~iterations, yielding $n=900$~total runs. The second uses DeepSeek-R1-0528-Qwen3-8B (8B parameters; recorded in the logs as \texttt{deepseek-r1}; hereafter DS8B) with the same DS70B Critic and QWQ Judge, again yielding $n=900$~runs. The third uses Claude Haiku~4.5~\citep{anthropic2024claude} (parameter count not public; hereafter Haiku) as the Actor and Claude Sonnet~4.6 (parameter count not public; hereafter Sonnet) as both Critic and primary Judge, with 5~repeats per persona--strategy cell on all three problems ($n=300$~runs each; 900 total).

For Haiku, all three problems now have full $n=300$ coverage. These runs
showed that Haiku converges on only ${\sim}1\%$ of Polchinski~2.7 cases under
the Sonnet Judge ($31\%$ under QWQ), leaving little dynamic range there for the Critic feedback strategy.
In what follows, we therefore report detailed Haiku Critic feedback strategy statistics
on the two Peskin~\&~Schroeder problems, where the loop is active, while still including Polchinski~2.7 in the per-problem QWQ convergence comparisons. With full coverage, cross-model comparisons involving Haiku are balanced, though its near-floor Polchinski~2.7 performance limits the Critic-strategy analysis there.

To separate effects rooted in the dialogue itself from those specific to a particular Judge (see \emph{e.g.}, \cite{tan2025judgebenchbenchmarkevaluatingllmbased} for a discussion on Judge reliability) we additionally re-score transcripts with independent Judges. For the Haiku runs, Sonnet is the primary Judge; we use closed-model Opus on a small sample as a calibration check on Sonnet/QWQ agreement. The Haiku transcripts are re-scored by QWQ and by DS70B, and the two DeepSeek-family Actor model settings are re-scored by DS70B alongside the QWQ primary Judge. Throughout the main results we report QWQ as the common scalable Judge across all three Actor model settings and treat DS70B re-scoring as a sensitivity check.

Since the tasks consist of classical theoretical physics problems which may be inspected manually, we review representative outlier transcripts and ambiguous Judge decisions, checking whether apparent failures reflect genuine physics errors, grading artifacts, or limitations of the model solution. The QWQ Judge produces scores that are broadly consistent with the manual review, and closer to the closed-model audit set than DS70B re-scoring, suggesting that the QWQ model may reliably act as the primary grader for the purposes of this study.

The analyzed corpus is compute-dense: across Actor, Critic, primary Judge, and re-scoring calls, it contains roughly $9.4\times10^7$ estimated token-equivalents reconstructed from archived prompt/response text, including about $6.2\times10^7$ tokens of judging. Exact provider-side token counters were not preserved, so these estimates are a scale-of-compute summary rather than billing totals. This complements broader single-turn studies such as TPBench, which spans 57 theoretical-physics problems and 10 models with five attempts per model/problem~\citep{chung2025tpbench}; SCALAR instead spends inference budget on factorial persona--strategy sweeps, iterative refinement, and Judge sensitivity over repeated transcripts. This comparison should not be read as a claim that unconstrained multi-agent dialogue is generically superior to single-turn evaluation: the SCALAR Critic is reference-conditioned, seeing the solution while being instructed not to disclose it, so the loop is closer to supervised tutoring than to open-ended autonomous discovery.
% ============================================================
\section{Results}
\label{sec:results}

We summarize performance using the evaluation metrics defined in~\cref{sec:eval}:
the mean per-turn score~$\bar{s}$, the gain~$g$, and the convergence rate~$R$.
When comparing strategies across problems, we additionally use the descriptive
problem-normalized contrasts $D_{\bar s}$ and $D_R$. We refer to one complete Actor--Critic--Judge dialogue trajectory as a \emph{run}; each run is
treated as one independent observation. For statistical comparisons, we use: Wilcoxon signed-rank to investigate the sign and significance of paired turn-$0$/turn-$(T{-}1)$ gain differences; Kruskal--Wallis to detect differences across the five
Critic feedback strategies; and Mann--Whitney for further pairwise Critic feedback strategy comparisons. These procedures use the raw run-level evaluation quantities, not the problem-normalized contrasts. Definitions and procedure details are given in \cref{app:metrics}.

\begin{figure}[t]
  \centering
  \includegraphics[width=\columnwidth]{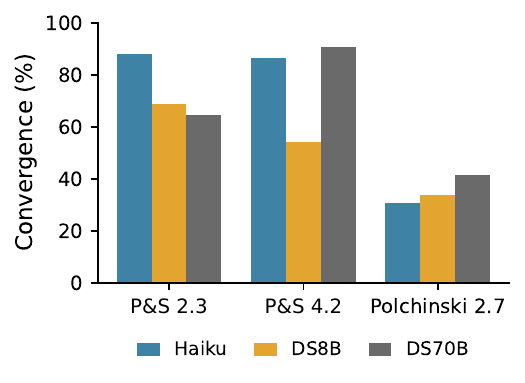}
  \caption{Per-problem convergence for the three Actor model settings. Haiku uses
    the Sonnet Critic; DS8B and DS70B both use the DS70B Critic.
    All shown cells use $n=300$ runs.
    }
  \label{fig:qwq_problem_convergence}
\end{figure}

\begin{figure}[t]
  \centering
  \includegraphics[width=\columnwidth]{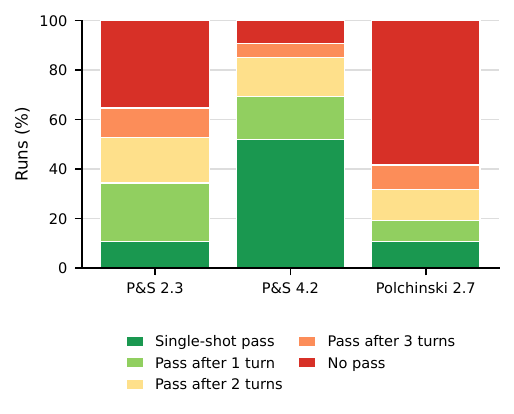}
  \caption{Run-fate breakdown by problem for the DS70B Actor under the
    QWQ Judge. Bars show single-shot passes, multi-turn passes after
    one/two/three Critic turns, and runs that terminate without passing.
    The large single-shot component on Peskin~\& Schroeder~4.2 reflects cases where the
    Actor passes before Critic feedback is needed.
    }
  \label{fig:per_problem}
\end{figure}

\paragraph{Structured feedback improves first attempts.}
For all three Actor model settings, the final solution is substantially
better than the Actor's first attempt under the common QWQ scoring
(\Cref{fig:qwq_problem_convergence}). On the two Peskin~\& Schroeder problems,
Haiku with a Sonnet Critic reaches high QWQ convergence ({$88.0\%$ on
Peskin~\& Schroeder~2.3 and $86.6\%$ on Peskin~\& Schroeder~4.2), while Polchinski~2.7
remains much harder---on the full} Polchinski~2.7
Haiku sweep, QWQ convergence is $30.9\%$ ($n=300$). The available
re-scoring checks preserve the sign of the multi-turn gain, so this
conclusion is not an artifact of the common QWQ scale.

For DS70B on all three problems, QWQ convergence reaches $65.7\%$
($n{=}900$), with mean final score $80.7$ and mean gain $g=+13.4$.
DS8B reaches lower QWQ convergence, $52.4\%$
($n{=}900$), with mean final score $76.6$ and mean gain $g=+13.3$.
This within-family comparison is useful because the Critic is held
fixed: DS8B resembles DS70B more
than Haiku in its Critic feedback strategy sensitivity, even though its absolute
performance is lower under QWQ. Under DS70B re-scoring the
within-family ordering becomes judge-sensitive, so the common figures
use QWQ rather than treating the DS8B/DS70B scale ordering as a
judge-independent claim.

The DS70B turn-$0$ score under QWQ averages $67.3$ and climbs to
$80.7$ by the end of the Critic loop, closing roughly $40\%$ of the gap
to saturation despite the stronger single-shot baseline. This
improvement is not a product of prompt optimization alone: it
requires the iterative structure, consistent with structured
multi-agent refinement reducing
errors~\citep{till2025multiconsistency} and contrasting with the
failure of naive sequential strategies~\citep{gao2025testtimescaling}
and with prior findings that single-turn prompt optimization yields
no significant gain~\citep{chung2025tpbench}.

The three problems span a clear difficulty spectrum for
DS70B, which manifests as a shift in the mixture of
run-fate categories (\Cref{fig:per_problem}). QWQ
convergence is $90.7\%$ on Peskin~\& Schroeder~4.2, $64.7\%$ on Peskin~\& Schroeder~2.3, and
$41.7\%$ on Polchinski~2.7, with $n=300$ runs per problem. DS8B does
not follow the same problem ordering: QWQ convergence is $69.0\%$ on
Peskin~\& Schroeder~2.3, $54.3\%$ on Peskin~\& Schroeder~4.2, and $34.0\%$ on Polchinski~2.7,
with $n=300$ runs per problem (\Cref{fig:qwq_problem_convergence}).
Polchinski~2.7 remains the hardest problem across all three Actor
model settings, but DS8B finds Peskin~\& Schroeder~2.3 easier than Peskin~\& Schroeder~4.2, unlike DS70B; under QWQ, Haiku converges at similar rates on the two Peskin~\& Schroeder problems ($88.0\%$ and $86.6\%$). 
The same reversal appears in the score-update curves discussed next:
for DS8B, Peskin~\& Schroeder~4.2 stops improving around intermediate scores,
whereas Peskin~\& Schroeder~2.3 continues to improve until higher scores.

\begin{figure*}[t]
  \centering
  \includegraphics[width=\textwidth]{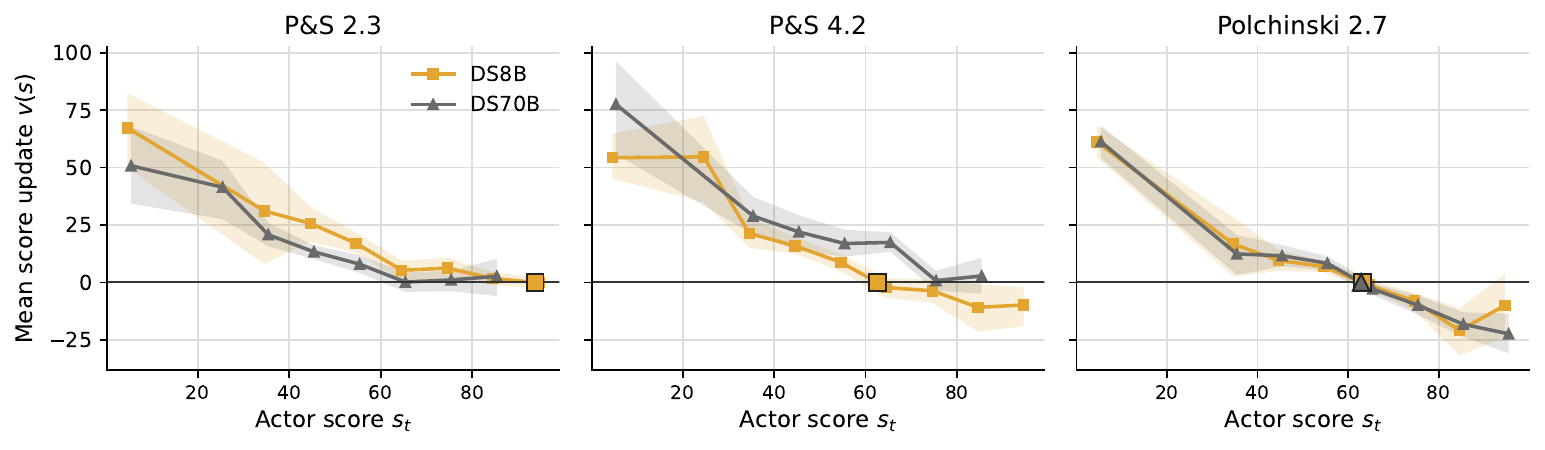}
  \caption{Empirical score-update curves for the two DeepSeek-family Actor
    scales under the common QWQ scoring. Points estimate the projected
    score-update field $v(s)=\mathbb{E}[\Delta s_t\mid s_t\simeq s]$ by averaging
    next-turn updates $\Delta s_t=s_{t+1}-s_t$ in bins of the current
    Actor score $s_t$ for the indicated Actor model and problem, pooling over
    personas and Critic feedback strategies; shaded bands are bootstrap $95\%$
    confidence intervals.
    Only runs with at least one observed update contribute, so
    immediate single-shot passes are not part of these curves. Markers
    on the zero line indicate estimated fixed points $s^\ast$, meaning
    zeros of the projected score-update field.}
  \label{fig:ds_param_scaling}
\end{figure*}

\paragraph{Score-update curves describe dialogue dynamics.}
The per-problem convergence rates say where the Actor--Critic loop
ends, but not how later Critic turns move the solution. Because SCALAR
records the fixed configuration and dialogue state after each turn
(\cref{sec:roles}), that recorded state is the natural Markov state
of the generated dialogue. The Markovian interpretation first becomes
interesting when we project that state to a single observable: the Judge
score. For the DeepSeek-family comparison in
\Cref{fig:ds_param_scaling}, we estimate the projected
\emph{score-update field}
\[
v(s)=\mathbb{E}\!\left[s_{t+1}-s_t\,\middle|\,s_t\simeq s\right].
\]
The average is over all observed next-turn transitions in the same score bin
for a fixed Actor model and problem, pooling over personas and Critic
strategies.
A zero of this curve is a \emph{projected fixed point}: the next Critic
turn no longer improves the Actor on average in the observed ensemble
near that score. Throughout the rest of the paper, ``fixed point'' refers
to this projected score-level object, not to a complete fixed point of
the full recorded dialogue state. This is the quantity that separates ``already
easy,'' ``still improvable,'' and ``apparently stuck'' regimes that are
compressed together by endpoint gain. In Appendix~\ref{app:dynamics},
the gain distributions show the same point from a different angle:
endpoint gain mixes first-shot successes, rescued runs, and stuck
trajectories rather than isolating a clean strategy effect.

Recent work has imported dynamical-systems language into LLM behavior:
\citet{carson2025statphys} models sentence-level reasoning trajectories
of open-source LLMs as a switching linear dynamical system on a latent
manifold, \citet{sarfati2025lines} characterize token-level activation
paths as approximately linear-drift stochastic processes, and
\citet{wang2025attractorcycles} study attractor cycles under repeated
paraphrasing. SCALAR puts related language to work in a different
setting: the underlying state is a multi-agent dialogue, the driving
field is supplied by a reference-conditioned Critic, and the observable
is an externally judged physics score. The result is a score projection
of a Markov transcript process. Appendix~\ref{app:dynamics}
states the corresponding projection and transition-kernel definitions,
and reports residual checks that support the score projection as a useful
coarse-grained diagnostic while leaving predictive kernel modeling to
larger studies.

Under this view, the within-family comparison reveals different regimes.
On Peskin~\& Schroeder~2.3 both DeepSeek-family Actors continue to improve until they reach high
scores. On Peskin~\& Schroeder~4.2, DS8B develops a mid-score fixed point near
$s_t\simeq 63$, whereas DS70B remains in a positive-drift regime through
most of the observed range and only crosses near high score; we
therefore do not mark a clean mid-score fixed point for DS70B on this
problem. On Polchinski~2.7, however, both DS8B and DS70B have a similar
fixed-point region near $s_t\simeq 63$. In this sense, increasing
parameter count within the same model family changes the easier-problem
dynamics but does not remove the hard-problem bottleneck.

\paragraph{Critic feedback strategy is model dependent.}

Feedback style should matter most when the Critic is actively moving the
Actor within a dialogue, rather than when a run passes immediately or remains
stuck. This is what we observe for Haiku, but not as a stable single-model
effect for either DeepSeek Actor scale. Thus the positive claim that constructive feedback helps is restricted to the asymmetric Haiku--Sonnet pairing; across same-family DeepSeek settings the better-supported general conclusion is the negative one that harsher feedback is not reliably best~\citep{kluger1996feedback,baron1988criticism}.

\begin{figure}[t]
  \centering
  \includegraphics[width=\columnwidth]{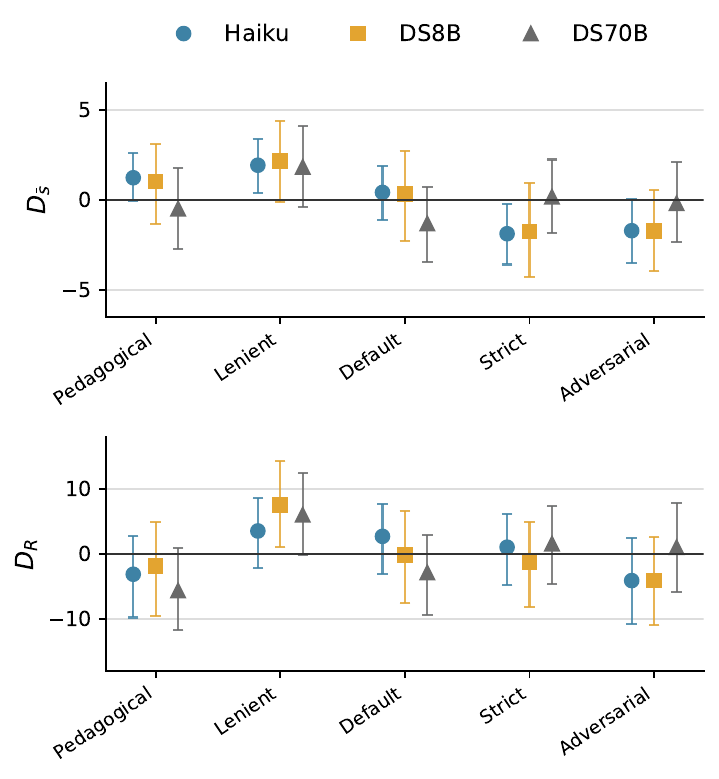}
\caption{Problem-normalized Critic feedback strategy contrasts under QWQ scoring.
    Haiku values use the two Peskin~\& Schroeder problems ($n{=}600$); DS8B and
    DS70B values use all three problems ($n{=}900$ each). For each
    Actor model setting and problem, we subtract the local
    baseline before averaging by Critic feedback strategy, so the figure asks
    which Critic feedback strategies sit above or below the corresponding
    model--problem baseline rather than which problems are easiest.
    Vertical bars show descriptive run-level bootstrap $95\%$ confidence
    intervals after problem-centering.
    $D_{\bar s}$ is measured in score points on the $0$--$100$ Judge
    scale; $D_R$ is measured in percentage points of convergence.}
  \label{fig:cross_system}
\end{figure}
Under the common QWQ scoring, Haiku is more sensitive to Critic
feedback strategy than either DeepSeek Actor scale (\Cref{fig:cross_system}).
The normalization is needed because raw mean score and raw convergence
are dominated by problem difficulty; the contrasts instead isolate
within-problem Critic feedback strategy shifts. Under this view, Haiku places pedagogical,
lenient, and default feedback above its local baseline in $D_{\bar s}$,
with strict and adversarial below it. This visual pattern matches the
raw $\bar s$ Critic feedback strategy test for Haiku (Kruskal--Wallis $p{=}0.0009$),
while the corresponding convergence-rate test is not significant
($p{=}0.27$). Thus the Haiku Critic feedback strategy signal is clearest in mean
per-turn score and should not be read as a strict ranking of the five
strategies.

For DS70B alone the five Critic feedback strategies are statistically
indistinguishable. Across the $900$-run sweep, the raw QWQ $\bar s$
means by Critic feedback strategy span only a few score points, and the omnibus
Critic-feedback-strategy tests are null
(Kruskal--Wallis $p{=}0.61$ for $\bar s$ and $p{=}0.17$ for $R$ under
QWQ). DS70B re-scoring gives the same null conclusion. DS8B is similar:
under QWQ, the omnibus tests find no reliable Critic feedback strategy effect on
$\bar s$ ($p{=}0.10$) or convergence rate~$R$ ($p{=}0.22$), and DS70B re-scoring agrees. Descriptively,
however, both
DeepSeek-family Actors put lenient feedback first under QWQ. Naively
pooling the two DeepSeek Actor scales yields a weak lenient advantage
for raw $\bar s$ and convergence rate. We therefore report this as a
QWQ-conditioned tendency, not as the same kind of robust Critic feedback strategy effect
seen for Haiku. Across all three Actor model settings, we find no stable
evidence that adversarial or strict feedback is the best Critic feedback strategy.

\paragraph{Actor persona prompting has negligible effect.}

Actor persona prompts are even weaker as a design variable. For DS70B under the common QWQ
scoring, the $12$ Actor-persona means span only
$\bar s \in [69.4, 74.1]$ on the full $900$-run sweep --- a $5$-point
range across $12$ configurations,
no larger than what sampling variation alone would produce if all
configurations had identical effects (Kruskal--Wallis $p{=}0.99$).
Varying the Actor persona along expertise
and reasoning-style axes has no measurable effect on DS70B outcomes.
DS8B gives the same broad message: the few
low-$p$ axis tests are not stable across evaluation metrics or
re-scoring checks, while the expertise axis itself is null.

Appendix~\ref{app:prompts} shows the full $60$-cell
persona$\times$strategy heatmaps for the balanced DeepSeek-family
sweeps, supporting the same conclusion: no single Actor persona is
consistently bright across problems, models, and Critic strategies.

For Haiku under QWQ scoring, the expertise axis also does not robustly
beat the unspecified default once we restrict to the two
Peskin~\&~Schroeder problems ($\bar s$: default~$76.7$ vs
expert~$76.4$). The available re-scoring checks do not turn this into a
stable dialogue-level expertise effect.
Taken together with the Critic feedback strategy analysis above, the lesson is
that among the prompt axes we tested, only Critic feedback strategy reliably
moves Haiku mean-score results, while neither DeepSeek Actor scale shows a robust
single-model persona or Critic feedback strategy effect. Actor persona prompting is
therefore not a reliable design variable in either pairing.

% ============================================================
\section{Discussion and Outlook}
\label{sec:discussion}

\subsection{Practical Takeaways}

Three observations from our experiments may be useful when
deploying SCALAR for graduate-level physics reasoning.

\emph{The Actor--Critic pairing is the key design choice.}
Critic feedback strategy matters, but not universally: it must be tuned for the
Actor--Critic system rather than chosen as a general rule. The
same multi-turn structure improves all three Actor model settings, but the
mechanism differs: Haiku improves smoothly within the dialogue, whereas
the DeepSeek-family averages combine first-shot successes, rescued runs,
and failures that remain stuck.
For a physicist using a frontier assistant, this gives three practical
lessons:
% \vspace{-0.05cm}
\begin{enumerate}
\item Use dialogue rather than only a one-shot request.\footnote{This
recommendation is hardly revolutionary, but it is often the missing step
in practice: a single flawed response is still sometimes treated by
physicists as a verdict on the model, rather than as the first turn of
an interaction.}
Across all
Actor model settings, multi-turn refinement improves over the initial
solution, especially when later turns target concrete physics checks
such as dimensions, limits, symmetries, known special cases, and
missing factors.
\item Do not expect Actor persona prompts to carry the result.
Pre-prompting current reasoning Actors with stronger personas, such as
``be an expert theorist,'' is not a reliable design variable in our
experiments.
\item Treat Critic feedback strategy as the prompt lever most worth testing, but do not assume
that harsher criticism is better. Strict and adversarial feedback are
never stably best in our experiments. Among the axes studied here, this
is where we see the clearest model-dependent movement in mean score. More precisely, we do not claim a universal constructive-over-adversarial ranking: the clear constructive signal appears in the asymmetric Haiku--Sonnet setting, while the DeepSeek results support only the weaker statement that strict/adversarial modes fail to dominate. A better starting point is
constructive feedback that preserves correct partial work, targets the
missing physics check, and only then tightens the standard~\citep{shute2008formative,hattie2007feedback,chi2001tutoring}; in the
DeepSeek-family settings, the lenient signal is suggestive but remains
QWQ-conditioned rather than a robust prescription.
\end{enumerate}
\vspace{-0.06cm}
Furthermore, recent results that
support the importance of asymmetric Actor--Critic pairing include
\cite{jiang2026asymmetricactorcriticmultiturnllm}. This also suggests
that prompt-engineering folklore deserves re-testing on current
reasoning-tuned models, especially in dialogue rather than only in
single-turn settings.

\emph{Dialogue dynamics are regime dependent.} Recent controlled studies
of reasoning models report complexity-dependent regimes, including
abrupt performance collapse beyond critical task difficulty and
phase-transition-like behavior in logical reasoning
benchmarks~\citep{shojaee2025illusionthinking,hazra2025threesatphase,zhang2026logicalphase}.
Our three physics problems are not a controlled complexity ladder, so
we do not claim a phase transition.

Instead, the score-update analysis in \cref{sec:results} gives a
conservative way to describe dialogue regimes: easy instances often pass
before Critic feedback, intermediate instances can be moved by structured
feedback, and hard instances can remain stuck despite additional
dialogue. The DS8B/DS70B comparison is a first controlled scale probe
within one model family: increasing
parameters changes the easy/intermediate regimes, but does not by itself
remove the Polchinski bottleneck. A denser DS-R1 scale ladder could test
whether such fixed-point regions disappear at problem-dependent critical
scales; with only two scales here, this remains motivation for future work
rather than a scaling claim.

In Appendix~\ref{app:scoring}, \cref{fig:app_score_components} also
identifies justification quality as the most persistent rubric gap. This
suggests a natural next step for SCALAR: treat Actor--Critic exchanges
as controlled dynamical systems, vary model scale, Critic strength, turn
budget, and problem difficulty, and test whether empirical transition
kernels predict held-out convergence. A richer state could also include
Critic scores, turning the one-dimensional Actor-score projection into a
joint description of feedback quality and Actor uptake.

This connects to the ``agentic gap''
critique of static text-only reasoning
evaluations~\citep{khan2025agenticgap}: scaffolding can rescue some
failures, but our hard-problem results show that scaffolding itself has
boundaries.

\subsection{Toward AI-Assisted Scientific Discovery}

AI-assisted discovery in fundamental physics is moving from speculation to an operational research question~\citep{lu2025languageagentsphysics}. Recent Gemini case studies give an open-ended perspective: frontier models can be useful in scientific workflows when embedded in human-supervised loops involving decomposition, critique, code generation, and external validation~\citep{woodruff2026accelerating}. SCALAR takes the complementary controlled direction: by returning to known-reference problems, it makes one part of that workflow measurable, namely how Critic feedback changes the trajectory of an Actor's reasoning. 

At the model level, one might expect that fine-tuning models on domain-specific high-energy theory corpora, as in the FeynTune study~\citep{Richmond:2025lzg}, would substantially improve research-style reasoning in high-energy theory. However, the results of that study indicate that targeted fine-tuning at the scales currently accessible to small academic collaborations does not close the gap with frontier general-purpose models. This motivates the SCALAR approach: rather than trying
to compete with frontier capability at the model level, scaffold a
frontier model with a structured Critic and obtain additional gains at
the \emph{interaction} level.
% Our benchmark problems are drawn from
% standard graduate-level references in quantum field
% theory and string
% theory; solving them reliably in
% collaboration with a Critic is a prerequisite --- though far from
% sufficient --- for contributing to research-level problems in the same
% area.  
We test this idea on graduate-level QFT and string-theory calculations in a controlled benchmark. Accordingly, ``scientific discovery'' is an aspirational target for future solution-free workflows and not a description of what the present experiments, where Critics have access to the solution, have already achieved.

However, in the present setup, the Critic is reference-conditioned: it has access to reference solutions and therefore guides the Actor with privileged knowledge of the answer. The Judge, by contrast, represents the external-validation part of the physicist's supervisory role: it evaluates the Actor--Critic exchange from outside the dialogue and determines whether the final reasoning meets the required standard. This is appropriate for benchmarking, but it means that the grading system still presupposes a known solution.

The longer-term vision is to remove this dependence on reference solutions. 
% More broadly, AI systems have begun to recover human-interpretable scientific structure from data~\citep{schmidt2009distilling,udrescu2020ai,cranmer2020discovering,kitouni2024neurons,richardson2025dna}, but turning such structure into scientific understanding often remains human-intensive.
\footnote{This connects to a broader program in AI-assisted science: models can already learn useful structure from scientific data, but extracting reliable human-usable insight from that structure often remains labor intensive~\citep{schmidt2009distilling,udrescu2020ai,cranmer2020discovering,kitouni2024neurons,richardson2025dna}. Agentic workflows could eventually help automate parts of this interpretive loop.}
For open problems, the researcher would instead move into the Critic role, supplying consistency checks and domain intuition. The Judge would then be replaced by an external arbiter --- experimental measurements, simulation outputs, theorem-prover checks, or mathematical consistency conditions --- so that the dialogue is validated against available ground truth rather than against a textbook solution. \Cref{app:scoring} explains which parts of the rubric would survive, and which would need replacement, in such a no-reference setting. In this setting, the three-agent analogy with the scientific method would become literal rather than stylized. Understanding how to configure Actor and Critic so that their dialogue reliably converges on problems whose answers are known is therefore a prerequisite for the more ambitious goal of collaborating on problems whose answers are not.

\section{Author contributions}
All authors contributed equally.

\section*{Data availability}
The \texttt{SCALAR} package, along with example problems, is available at \url{https://github.com/xand-stapleton/ai_agents}.

\section*{Acknowledgments}
CP was partially supported by the Science and Technology Facilities Council (STFC) Consolidated Grant ST/X00063X/1 `Amplitudes, Strings \& Duality.' AGS acknowledges support from Pierre Andurand. This research utilized the Apocrita HPC facility, supported by QMUL Research-IT \cite{king_2017_438045}. ST is supported by the Swiss National Science Foundation project number P5R5PT\_222350, and acknowledges the CERN TH Department for hospitality while this research was being carried out.

% ============================================================
\bibliography{references}

@book{peskin1995introduction,
  title={An Introduction to Quantum Field Theory},
  author={Peskin, Michael E. and Schroeder, Daniel V.},
  year={1995},
  publisher={Addison-Wesley}
}

@book{polchinski1998string,
  title={String Theory},
  author={Polchinski, Joseph},
  volume={1},
  year={1998},
  publisher={Cambridge University Press}
}

@book{vygotsky1978,
  author    = {Vygotsky, Lev Semyonovich},
  title     = {Mind in Society: The Development of Higher Psychological Processes},
  year      = {1978},
  publisher = {Harvard University Press},
  address   = {Cambridge, MA},
  note      = {Edited by Michael Cole, Vera John-Steiner, Sylvia Scribner, and Ellen Souberman}
}

@article{wood1976scaffolding,
  author    = {Wood, David and Bruner, Jerome S. and Ross, Gail},
  title     = {The Role of Tutoring in Problem Solving},
  journal   = {Journal of Child Psychology and Psychiatry},
  volume    = {17},
  number    = {2},
  pages     = {89--100},
  year      = {1976},
  doi       = {10.1111/j.1469-7610.1976.tb00381.x}
}

@article{wilcoxon1945individual,
  author  = {Wilcoxon, Frank},
  title   = {Individual Comparisons by Ranking Methods},
  journal = {Biometrics Bulletin},
  volume  = {1},
  number  = {6},
  pages   = {80--83},
  year    = {1945},
  doi     = {10.2307/3001968}
}

@article{mann1947test,
  author  = {Mann, Henry B. and Whitney, Donald R.},
  title   = {On a Test of Whether One of Two Random Variables is Stochastically Larger than the Other},
  journal = {The Annals of Mathematical Statistics},
  volume  = {18},
  number  = {1},
  pages   = {50--60},
  year    = {1947},
  doi     = {10.1214/aoms/1177730491}
}

@article{kruskal1952use,
  author  = {Kruskal, William H. and Wallis, W. Allen},
  title   = {Use of Ranks in One-Criterion Variance Analysis},
  journal = {Journal of the American Statistical Association},
  volume  = {47},
  number  = {260},
  pages   = {583--621},
  year    = {1952},
  doi     = {10.1080/01621459.1952.10483441}
}

@article{deepseek2025r1,
  title={DeepSeek-R1: Incentivizing Reasoning Capability in LLMs via Reinforcement Learning},
  author={{DeepSeek-AI}},
  journal={arXiv preprint arXiv:2501.12948},
  year={2025}
}

@misc{anthropic2024claude,
  title={{The Claude Model Card and System Prompt}},
  author={{Anthropic}},
  year={2024},
  note={\url{https://www.anthropic.com}}
}

@article{romera2024mathematical,
  title={Mathematical Discoveries from Program Search with Large Language Models},
  author={Romera-Paredes, Bernardino and others},
  journal={Nature},
  volume={625},
  pages={468--475},
  year={2024}
}

@article{du2024improving,
  title={Improving Factuality and Reasoning in Language Models through Multiagent Debate},
  author={Du, Yilun and Li, Shuang and Torralba, Antonio and Tenenbaum, Joshua B. and Mordatch, Igor},
  journal={arXiv preprint arXiv:2305.14325},
  year={2024}
}

@misc{qwq2024,
  title={{QwQ: Reflect Deeply on the Boundaries of the Unknown}},
  author={{Qwen Team}},
  year={2024},
  note={\url{https://qwenlm.github.io/blog/qwq-32b-preview/}}
}

@article{guevara2026singleminus,
  title={Single-minus gluon tree amplitudes are nonzero},
  author={Guevara, Alfredo and Lupsasca, Alexandru and Skinner, David and Strominger, Andrew and Weil, Kevin},
  journal={arXiv preprint arXiv:2602.12176},
  year={2026}
}

@article{chung2025tpbench,
  title={Theoretical Physics Benchmark ({TPBench}): A Dataset and Study of {AI} Reasoning Capabilities in Theoretical Physics},
  author={Chung, Daniel and others},
  journal={arXiv preprint arXiv:2502.15815},
  year={2025}
}

@article{lu2025languageagentsphysics,
  title={Can Theoretical Physics Research Benefit from Language Agents?},
  author={Lu, Sirui and Jin, Zhijing and Zhang, Terry Jingchen and Kos, Pavel and Cirac, J. Ignacio and Sch{\"o}lkopf, Bernhard},
  journal={arXiv preprint arXiv:2506.06214},
  year={2025}
}

@article{shojaee2025illusionthinking,
  title={The Illusion of Thinking: Understanding the Strengths and Limitations of Reasoning Models via the Lens of Problem Complexity},
  author={Shojaee, Parshin and Mirzadeh, Iman and Alizadeh, Keivan and Horton, Maxwell and Bengio, Samy and Farajtabar, Mehrdad},
  journal={arXiv preprint arXiv:2506.06941},
  year={2025}
}

@article{khan2025agenticgap,
  title={A Comment On ``The Illusion of Thinking'': Reframing the Reasoning Cliff as an Agentic Gap},
  author={Khan, Sheraz and Madhavan, Subha and Natarajan, Kannan},
  journal={arXiv preprint arXiv:2506.18957},
  year={2025}
}

@article{hazra2025threesatphase,
  title={Have Large Language Models Learned to Reason? A Characterization via 3-{SAT} Phase Transition},
  author={Hazra, Rishi and Venturato, Gabriele and Zuidberg Dos Martires, Pedro and De Raedt, Luc},
  journal={arXiv preprint arXiv:2504.03930},
  year={2025}
}

@article{zhang2026logicalphase,
  title={Logical Phase Transitions: Understanding Collapse in {LLM} Logical Reasoning},
  author={Zhang, Xinglang and Zhang, Yunyao and Chen, ZeLiang and Yu, Junqing and Yang, Wei and Song, Zikai},
  journal={arXiv preprint arXiv:2601.02902},
  year={2026}
}

@article{Novikov:2025alphaevolve,
  title={{AlphaEvolve}: A Coding Agent for Scientific and Algorithmic Discovery},
  author={Novikov, Alexander and Vu, Ngan and Eisenberger, Marvin and Dupont, Emilien and Huang, Po-Sen and Wagner, Adam Zsolt and Shirobokov, Sergey and Kozlovskii, Borislav and Ruiz, Francisco J. R. and Mehrabian, Abbas and Kumar, M. Pawan and See, Abigail and Chaudhuri, Swarat and Holland, George and Davies, Alex and Nowozin, Sebastian and Kohli, Pushmeet and Balog, Matej},
  journal={arXiv preprint arXiv:2506.13131},
  year={2025}
}

@article{Plehn:2026gxv,
  title={{MadAgents}},
  author={Plehn, Tilman and Schiller, Daniel and Schmal, Nikita},
  journal={arXiv preprint arXiv:2601.21015},
  year={2026}
}

@article{Agrawal:2026lvg,
  title={{The FERMIACC}: Agents for Particle Theory},
  author={Agrawal, Prateek and Craig, Nathaniel and Madden, Amalia and Valenzuela Lombera, Inigo},
  journal={arXiv preprint arXiv:2603.22538},
  year={2026}
}

@inproceedings{hwang2025instructionaldistractions,
  title={{LLM}s can be easily Confused by Instructional Distractions},
  author={Hwang, Yerin and Kim, Yongil and Koo, Jahyun and Kang, Taegwan and Bae, Hyunkyung and Jung, Kyomin},
  booktitle={Proceedings of the 63rd Annual Meeting of the Association for Computational Linguistics (Volume 1: Long Papers)},
  pages={19483--19496},
  year={2025},
  publisher={Association for Computational Linguistics},
  doi={10.18653/v1/2025.acl-long.957},
  url={https://aclanthology.org/2025.acl-long.957/}
}

@article{gao2025testtimescaling,
  title={Test-time Scaling Techniques in Theoretical Physics: A Comparison of Methods on the {TPBench} Dataset},
  author={Gao, Zhiyuan and others},
  journal={arXiv preprint},
  year={2025}
}

@article{jaiswal2024mora,
  title={Improving Physics Reasoning in Large Language Models Using Mixture of Refinement Agents},
  author={Jaiswal, Raj and others},
  journal={arXiv preprint arXiv:2412.00821},
  year={2024}
}

@article{laban2025lost,
  title={{LLMs} Get Lost In Multi-Turn Conversation},
  author={Laban, Philippe and others},
  journal={arXiv preprint},
  year={2025}
}

@article{estornell2024accdebate,
  title={{ACC-Collab}: An Actor-Critic Approach to Multi-Agent {LLM} Collaboration},
  author={Estornell, Andrew and Ton, Jean-Francois and Yao, Yuanshun and Liu, Yang},
  journal={arXiv preprint arXiv:2411.00053},
  year={2024}
}

@article{xu2025multiagentphysicist,
  title={Advancing {AI}-Scientist Understanding: Multi-Agent {LLMs} with Interpretable Physics Reasoning},
  author={Xu, Yinggan and Kimlee, Hana and Xiao, Yijia and Luo, Di},
  journal={arXiv preprint arXiv:2504.01911},
  year={2025},
  note={ICML 2025 Workshop on MAS}
}

@article{gupta2024persona,
  title={Persona is a Double-Edged Sword: Enhancing the Zero-shot Reasoning by Ensembling the Role-playing and Neutral Prompts},
  author={Gupta, Junseok and others},
  journal={arXiv preprint arXiv:2408.08631},
  year={2024}
}

@article{zhang2025turnbench,
  title={{TurnBench-MS}: A Benchmark for Evaluating Multi-Turn, Multi-Step Reasoning in Large Language Models},
  author={Zhang, Yifan and others},
  journal={arXiv preprint},
  year={2025}
}

@article{carson2025statphys,
  title={A Statistical Physics of Language Model Reasoning},
  author={Carson, Timothy},
  journal={arXiv preprint},
  year={2025}
}

@inproceedings{sarfati2025lines,
  title={Lines of Thought in Large Language Models},
  author={Sarfati, Raphael and others},
  booktitle={ICLR},
  year={2025}
}

@article{zhang2025physreason,
  title={{PhysReason}: A Comprehensive Benchmark towards Physics-Based Reasoning},
  author={Zhang, Xinyu and Dong, Yiwen and others},
  journal={arXiv preprint arXiv:2502.12054},
  year={2025}
}

@article{wang2025attractorcycles,
  title={Unveiling Attractor Cycles in Large Language Models: A Dynamical Systems View of Successive Paraphrasing},
  author={Wang, Yibo and others},
  journal={arXiv preprint},
  year={2025}
}

@article{till2025multiconsistency,
  title={Multi-Model Consistency Improves Hallucination Detection and Mitigation in Large Language Models},
  author={Till, Robert and others},
  journal={arXiv preprint},
  year={2025}
}

@article{Shih:2026jfe,
    author = "Shih, David",
    title = "{Learning to Unscramble Feynman Loop Integrals with SAILIR}",
    journal = "arXiv preprint arXiv:2604.05034",
    eprint = "2604.05034",
    archivePrefix = "arXiv",
    primaryClass = "hep-ph",
    month = "4",
    year = "2026"
}

@article{Richmond:2025lzg,
    author = "Richmond, Paul and Papageorgakis, Constantinos and Niarchos, Vasilis and Chowdhury, Borun and Agarwal, Prarit",
    title = {{FeynTune: large language models for high-energy theory}},
    eprint = "2508.03716",
    archivePrefix = "arXiv",
    primaryClass = "cs.CL",
    doi = "10.1088/2632-2153/ae47bb",
    journal = "Mach. Learn. Sci. Tech.",
    volume = "7",
    number = "2",
    pages = "025012",
    year = "2026"
}

@article{Shih:2026lmy,
    author = "Shih, David",
    title = "{Learning to Unscramble: Simplifying Symbolic Expressions via Self-Supervised Oracle Trajectories}",
    journal = "arXiv preprint arXiv:2603.11164",
    eprint = "2603.11164",
    archivePrefix = "arXiv",
    primaryClass = "hep-th",
    month = "3",
    year = "2026"
}

@article{Schwartz:2026ekw,
    author = "Schwartz, Matthew D.",
    title = "{Resummation of the C-Parameter Sudakov Shoulder Using Effective Field Theory}",
    journal = "arXiv preprint arXiv:2601.02484",
    eprint = "2601.02484",
    archivePrefix = "arXiv",
    primaryClass = "hep-ph",
    month = "1",
    year = "2026"
}

@manual{king_2017_438045,
  title        = {Apocrita - High Performance Computing Cluster for
                   Queen Mary University of London
                  },
  author       = {King, Thomas and
                  Butcher, Simon and
                  Zalewski, Lukasz},
  month        = mar,
  year         = 2017,
  doi          = {10.5281/zenodo.438045},
  url          = {https://doi.org/10.5281/zenodo.438045},
}

@misc{liang2024mathchatbenchmarkingmathematicalreasoning,
      title={MathChat: Benchmarking Mathematical Reasoning and Instruction Following in Multi-Turn Interactions}, 
      author={Zhenwen Liang and Dian Yu and Wenhao Yu and Wenlin Yao and Zhihan Zhang and Xiangliang Zhang and Dong Yu},
      year={2024},
      eprint={2405.19444},
      archivePrefix={arXiv},
      primaryClass={cs.AI},
      url={https://arxiv.org/abs/2405.19444}, 
}

@misc{tan2025judgebenchbenchmarkevaluatingllmbased,
      title={JudgeBench: A Benchmark for Evaluating LLM-based Judges}, 
      author={Sijun Tan and Siyuan Zhuang and Kyle Montgomery and William Y. Tang and Alejandro Cuadron and Chenguang Wang and Raluca Ada Popa and Ion Stoica},
      year={2025},
      eprint={2410.12784},
      archivePrefix={arXiv},
      primaryClass={cs.AI},
      url={https://arxiv.org/abs/2410.12784}, 
}

@misc{jiang2026asymmetricactorcriticmultiturnllm,
      title={Asymmetric Actor-Critic for Multi-turn LLM Agents}, 
      author={Shuli Jiang and Zhaoyang Zhang and Yi Zhang and Shuo Yang and Wei Xia and Stefano Soatto},
      year={2026},
      eprint={2604.00304},
      archivePrefix={arXiv},
      primaryClass={cs.CL},
      url={https://arxiv.org/abs/2604.00304}, 
}

@inproceedings{NEURIPS2023_91f18a12,
 author = {Zheng, Lianmin and Chiang, Wei-Lin and Sheng, Ying and Zhuang, Siyuan and Wu, Zhanghao and Zhuang, Yonghao and Lin, Zi and Li, Zhuohan and Li, Dacheng and Xing, Eric and Zhang, Hao and Gonzalez, Joseph E and Stoica, Ion},
 booktitle = {Advances in Neural Information Processing Systems},
 editor = {A. Oh and T. Naumann and A. Globerson and K. Saenko and M. Hardt and S. Levine},
 pages = {46595--46623},
 publisher = {Curran Associates, Inc.},
 title = {Judging LLM-as-a-Judge with MT-Bench and Chatbot Arena},
 url = {https://proceedings.neurips.cc/paper_files/paper/2023/file/91f18a1287b398d378ef22505bf41832-Paper-Datasets_and_Benchmarks.pdf},
 volume = {36},
 year = {2023}
}

@article{woodruff2026accelerating,
  title={Accelerating Scientific Research with Gemini: Case Studies and Common Techniques},
  author={Woodruff, David P. and Cohen-Addad, Vincent and Jain, Lalit and Mao, Jieming and Zuo, Song and others},
  journal={arXiv preprint arXiv:2602.03837},
  year={2026},
  url={https://arxiv.org/abs/2602.03837}
}

@article{richardson2025dna,
  title={The DNA of nuclear models: How AI predicts nuclear masses},
  author={Richardson, Kate A. and Trifinopoulos, Sokratis and Williams, Mike},
  journal={arXiv preprint arXiv:2508.08370},
  year={2025},
  url={https://arxiv.org/abs/2508.08370}
}

@inproceedings{kitouni2024neurons,
  title={From Neurons to Neutrons: A Case Study in Interpretability},
  author={Kitouni, Ouail and Nolte, Niklas and P{\'e}rez-D{\'i}az, V{\'i}ctor Samuel and Trifinopoulos, Sokratis and Williams, Mike},
  booktitle={Proceedings of the 41st International Conference on Machine Learning},
  pages={24726--24748},
  year={2024},
  organization={PMLR},
  url={https://proceedings.mlr.press/v235/kitouni24a.html}
}

@article{schmidt2009distilling,
  title={Distilling free-form natural laws from experimental data},
  author={Schmidt, Michael and Lipson, Hod},
  journal={Science},
  volume={324},
  number={5923},
  pages={81--85},
  year={2009}
}

@article{udrescu2020ai,
  title={AI Feynman: A physics-inspired method for symbolic regression},
  author={Udrescu, Silviu-Marian and Tegmark, Max},
  journal={Science Advances},
  volume={6},
  number={16},
  pages={eaay2631},
  year={2020}
}

@inproceedings{cranmer2020discovering,
  title={Discovering symbolic models from deep learning with inductive biases},
  author={Cranmer, Miles and Sanchez-Gonzalez, Alvaro and Battaglia, Peter and Xu, Rui and Cranmer, Kyle and Spergel, David and Ho, Shirley},
  booktitle={Advances in Neural Information Processing Systems},
  volume={33},
  pages={17429--17442},
  year={2020}
}

@inproceedings{amershi2019guidelines,
author = {Amershi, Saleema and Weld, Dan and Vorvoreanu, Mihaela and Fourney, Adam and Nushi, Besmira and Collisson, Penny and Suh, Jina and Iqbal, Shamsi and Bennett, Paul N. and Inkpen, Kori and Teevan, Jaime and Kikin-Gil, Ruth and Horvitz, Eric},
title = {Guidelines for Human-AI Interaction},
year = {2019},
isbn = {9781450359702},
publisher = {Association for Computing Machinery},
address = {New York, NY, USA},
url = {https://doi.org/10.1145/3290605.3300233},
doi = {10.1145/3290605.3300233},
booktitle = {Proceedings of the 2019 CHI Conference on Human Factors in Computing Systems},
pages = {1–13},
numpages = {13},
location = {Glasgow, Scotland Uk},
series = {CHI '19}
}

@article{shneiderman2020human,
author = {Ben Shneiderman},
title = {Human-Centered Artificial Intelligence: Reliable, Safe \& Trustworthy},
journal = {International Journal of Human–Computer Interaction},
volume = {36},
number = {6},
pages = {495--504},
year = {2020},
publisher = {Taylor \& Francis},
doi = {10.1080/10447318.2020.1741118},
URL = {https://doi.org/10.1080/10447318.2020.1741118},
eprint = {https://doi.org/10.1080/10447318.2020.1741118}
}

@article{kluger1996feedback,
  author  = {Kluger, Avraham N. and DeNisi, Angelo},
  title   = {The Effects of Feedback Interventions on Performance:
             A Historical Review, a Meta-Analysis, and a Preliminary
             Feedback Intervention Theory},
  journal = {Psychological Bulletin},
  year    = {1996},
  volume  = {119},
  number  = {2},
  pages   = {254--284},
  doi     = {10.1037/0033-2909.119.2.254}
}

@article{baron1988criticism,
  author  = {Baron, Robert A.},
  title   = {Negative Effects of Destructive Criticism:
             Impact on Conflict, Self-Efficacy, and Task Performance},
  journal = {Journal of Applied Psychology},
  year    = {1988},
  volume  = {73},
  number  = {2},
  pages   = {199--207},
  doi     = {10.1037/0021-9010.73.2.199}
}

@article{shute2008formative,
  author  = {Shute, Valerie J.},
  title   = {Focus on Formative Feedback},
  journal = {Review of Educational Research},
  year    = {2008},
  volume  = {78},
  number  = {1},
  pages   = {153--189},
  doi     = {10.3102/0034654307313795}
}

@article{hattie2007feedback,
  author  = {Hattie, John and Timperley, Helen},
  title   = {The Power of Feedback},
  journal = {Review of Educational Research},
  year    = {2007},
  volume  = {77},
  number  = {1},
  pages   = {81--112},
  doi     = {10.3102/003465430298487}
}

@article{butler1987evaluation,
  author  = {Butler, Ruth},
  title   = {Task-Involving and Ego-Involving Properties of Evaluation:
             Effects of Different Feedback Conditions on Motivational
             Perceptions, Interest, and Performance},
  journal = {Journal of Educational Psychology},
  year    = {1987},
  volume  = {79},
  number  = {4},
  pages   = {474--482},
  doi     = {10.1037/0022-0663.79.4.474}
}

@article{chi2001tutoring,
  author  = {Chi, Michelene T. H. and Siler, Stephanie A. and
             Jeong, Heisawn and Yamauchi, Takashi and
             Hausmann, Robert G.},
  title   = {Learning from Human Tutoring},
  journal = {Cognitive Science},
  year    = {2001},
  volume  = {25},
  number  = {4},
  pages   = {471--533},
  doi     = {10.1207/S15516709COG2504_1}
}
\bibliographystyle{icml2026}

% ============================================================
\appendix

\section{Prompt Texts}
\label{app:prompts}

In SCALAR, Actor persona pre-prompting is factored into expertise
$\times$ reasoning style.
Representative summaries of each component follow; the full prompts
used in experiments are provided in the code repository.

\paragraph{Expertise level.}
\emph{Expert}: ``You are an expert in theoretical physics with deep
knowledge of QFT and mathematical methods.''
\emph{Novice}: ``You are a graduate student learning QFT, working through
problems to build understanding.''
\emph{Default~(d)}: no expertise instruction.

\paragraph{Reasoning style.}
\emph{Meticulous}: ``Check every algebraic step carefully. Verify signs,
prefactors, and limits before proceeding.''
\emph{Physical}: ``Prioritize physical intuition. Use dimensional analysis,
limiting cases, and symmetry to guide your calculation.''
\emph{Skeptical}: ``Question every assumption. If a step seems unjustified,
flag it and consider alternatives.''
\emph{Default~(d)}: no style instruction.

\paragraph{Feedback strategies.}
\emph{Adversarial}: ``Aggressively challenge every claim. Demand explicit
justification for each step.''
\emph{Strict}: ``Flag every error precisely with the correct form.''
\emph{Pedagogical}: ``Guide through Socratic questioning. Ask leading questions
rather than stating errors directly.''
\emph{Lenient}: ``Focus on what the solver got right. Offer gentle suggestions.
Accept partial progress.''
\emph{Default~(d)}: no additional style instructions.

\begin{figure*}[p]
  \centering
  \includegraphics[width=\textwidth]{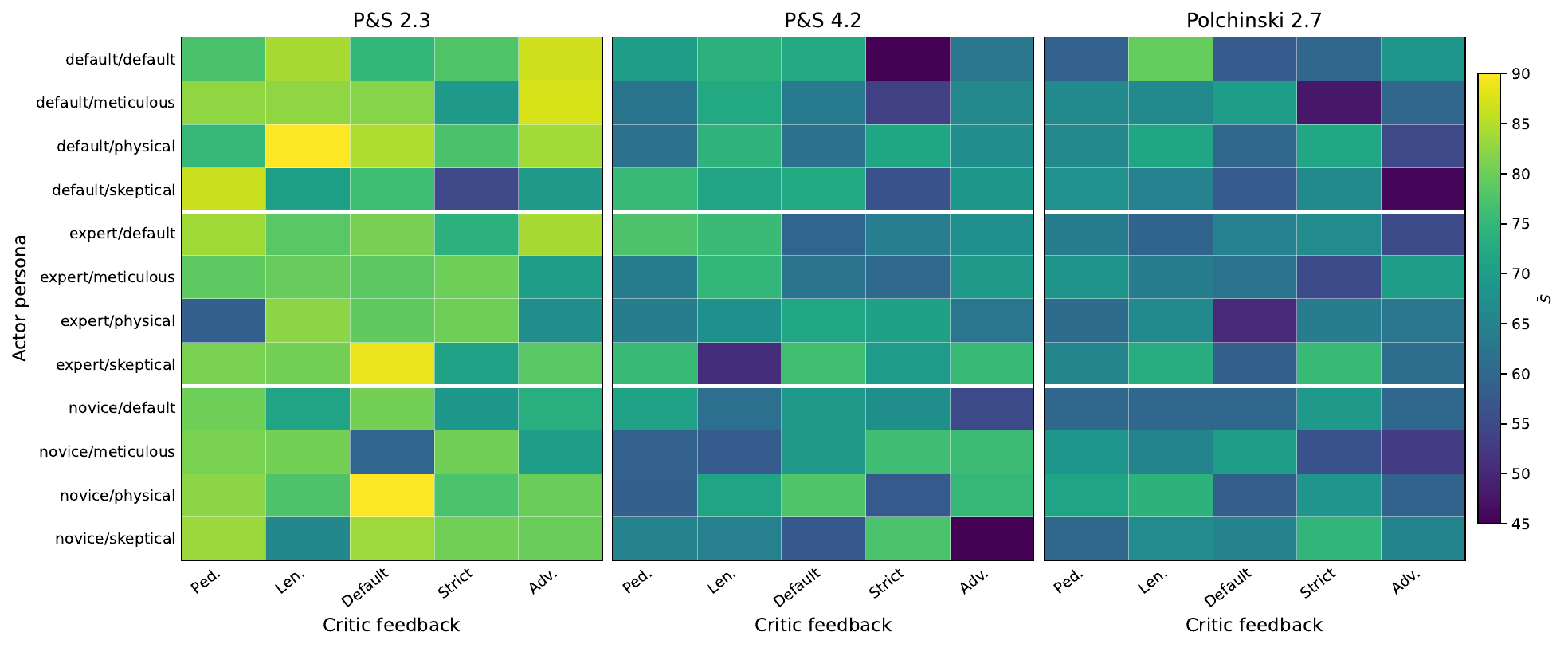}
  \caption{DS8B $\bar s$ heatmaps under QWQ scoring, separated by
    problem. Rows are Actor personas, columns are Critic feedback
    strategies, and cell color gives the mean per-turn score.}
  \label{fig:app_sbar_ds8b}
\end{figure*}

\begin{figure*}[p]
  \centering
  \includegraphics[width=\textwidth]{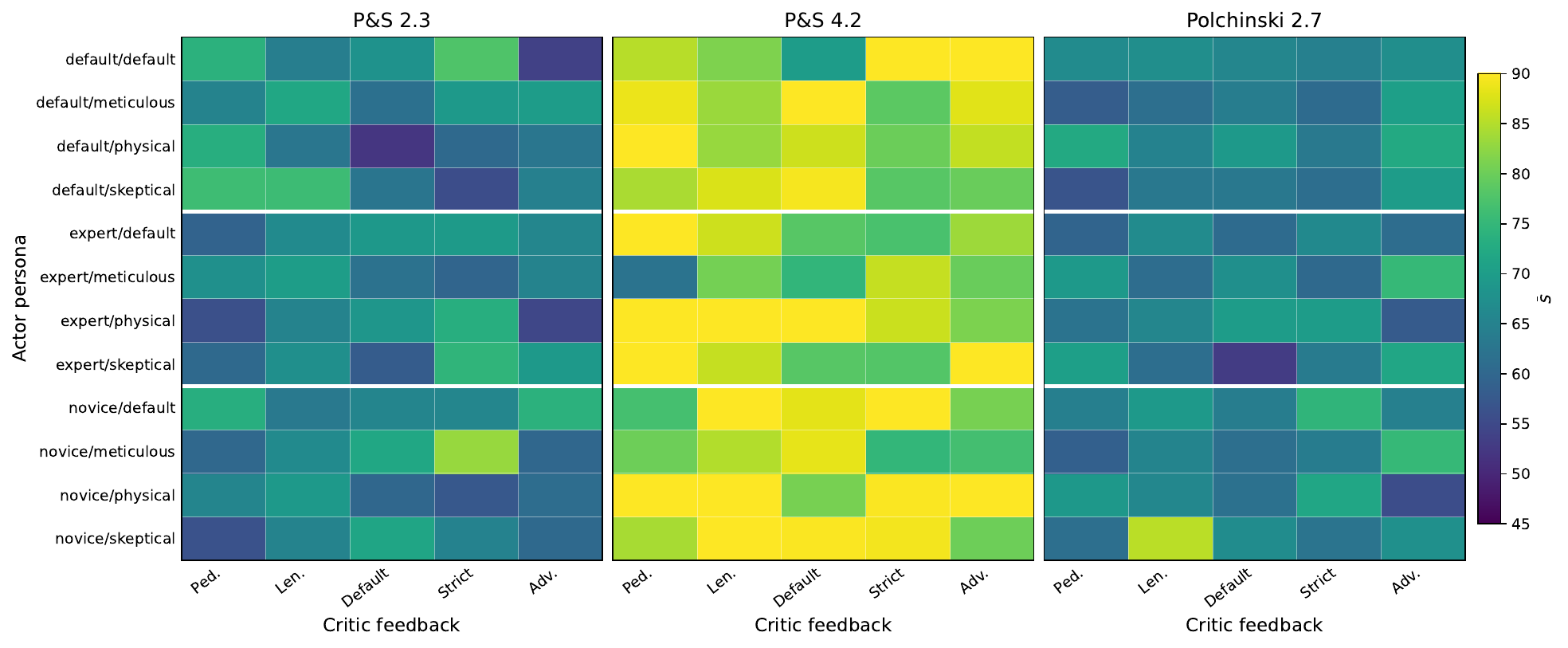}
  \caption{DS70B $\bar s$ heatmaps under QWQ scoring, separated by
    problem. Rows are Actor personas, columns are Critic feedback
    strategies, and cell color gives the mean per-turn score.}
  \label{fig:app_sbar_ds70b}
\end{figure*}

The heatmaps in \Cref{fig:app_sbar_ds8b,fig:app_sbar_ds70b} show the
balanced persona$\times$strategy grids for the two DeepSeek-family
sweeps. The block structure separates default, expert, and novice Actor
personas. The visual pattern is irregular rather than block-diagonal:
individual cells can be bright, but there is no stable persona formula
that dominates across problems.

\section{List of Physics Problems}
\label{app:probs}
As was outlined in \cref{sec:models}, in this work we consider three problems chosen to probe various qualities of each model-prompt pair. Each problem is reproduced below.

\paragraph{Peskin~\& Schroeder 2.3 \cite{peskin1995introduction}}
Evaluate the function
$$
\langle 0 | \phi(x)\phi(y) |0\rangle = D(x-y) = \int \frac{d^3 p}{(2\pi)^3}\frac{1}{E_{\vec p}}e^{-i p\cdot (x-y)}
$$
for $(x-y)$ spacelike so that $(x-y)^2=-r^2$, explicitly in terms of Bessel functions.

\paragraph{Polchinski 2.7 \cite{polchinski1998string}}
Consider the free CFT of free scalars $X^\mu$.
\begin{enumerate}
    \item By computing the relevant OPEs, confirm the following weights
    \begin{eqnarray*}
    && X^\mu ~~ (0,0)\\
    && \partial X^\mu ~~ (1,0)\\
    && \bar\partial X^\mu ~~(0,1)\\
    && \partial^2 X^\mu ~~ (2,0)\\
    && :e^{ik\cdot X}:~~ \left( \frac{\alpha' k^2}{4}, \frac{\alpha' k^2}{4}\right)
    \end{eqnarray*}   
    and determine which operators are tensors.
    \item Do this for the same operators in the linear dilaton theory.
\end{enumerate}

\paragraph{Peskin~\& Schroeder 4.2 \cite{peskin1995introduction}}
Consider the following Lagrangian, involving two real scalar fields $\Phi$ and $\phi$:
\[
\mathcal{L} = \tfrac{1}{2}(\partial_\mu \Phi)^2 - \tfrac{1}{2}M^2 \Phi^2 + \tfrac{1}{2}(\partial_\mu \phi)^2 - \tfrac{1}{2}m^2 \phi^2 - \mu \Phi \phi \phi.
\]
The last term is an interaction that allows a $\Phi$ particle to decay into two $\phi$'s, provided that $M > 2m$. Assuming that this condition is met, calculate the lifetime of the $\Phi$ to lowest order in $\mu$.

\section{Scoring Criteria}
\label{app:scoring}

We evaluate each candidate solution using an \emph{LLM Judge} that compares the submitted solution against the problem statement and, when available, a reference solution. The Judge produces a structured verdict consisting of binary pass/fail, fine-grained sub-scores, error flags, and qualitative summaries. The numerical thresholds below are operational grading conventions chosen for reproducibility and stopping decisions; larger follow-up studies should examine how the choice of grading criteria affects the results.

\paragraph{Inputs.}
For each evaluation instance, the Judge receives:
\begin{enumerate}
    \item the problem statement $P$,
    \item the Actor-generated solution $A$,
    \item Critic feedback $C$ embedded in the evaluated text, and
    \item a reference solution $R$ (when available).
\end{enumerate}

\paragraph{Final-result equivalence check.}
Before assigning scores, the Judge performs an explicit \emph{final-result comparison} between the Actor's claimed final answer and the reference answer. Let
\[
\mathrm{Eq}(A, R) \in \{0,1\}
\]
denote whether the Actor's final result is mathematically equivalent to the reference. Equivalence is defined up to:
\begin{itemize}
    \item algebraic rearrangement,
    \item equivalent notation,
    \item trivial reordering of factors.
\end{itemize}
Non-equivalence is declared if the two results differ in sign, numerical prefactor, functional form, missing or extra terms, or complex phase factors. If the Actor provides no final formula or only an incomplete formula, then
\[
\mathrm{Eq}(A,R)=0.
\]

\paragraph{Actor score.}
The Actor is evaluated along six dimensions:
\[
S_{\mathrm{actor}}
= S_c + S_r + S_l + S_j + S_m + S_p,
\qquad
S_{\mathrm{actor}} \in [0,100],
\]
where:
\begin{align*}
S_c &\in [0,50] && \text{correctness},\\
S_r &\in [0,10] && \text{mathematical rigor},\\
S_l &\in [0,10] && \text{logical flow},\\
S_j &\in [0,10] && \text{quality of justification},\\
S_m &\in [0,10] && \text{completeness},\\
S_p &\in [0,10] && \text{physical consistency}.
\end{align*}

\subparagraph{Correctness ($S_c$).}
Correctness is based on the error density
\[
\rho = \frac{N_{\mathrm{errors}}}{N_{\mathrm{steps}}}.
\]
The rubric is:
\begin{itemize}
    \item $S_c = 50$: $\rho = 0$, all equations are mathematically sound, and the final result is within $1\%$ of the expected result;
    \item $S_c \in [42,49]$: $\rho < 0.1$, only minor computational errors;
    \item $S_c \in [27,41]$: $0.1 \le \rho < 0.3$, some correct intermediate reasoning but substantial gaps;
    \item $S_c \in [0,26]$: $\rho \ge 0.3$, indicating fundamental conceptual or methodological errors.
\end{itemize}

\subparagraph{Mathematical rigor ($S_r$).}
Rigor is measured by the justification ratio
\[
J = \frac{N_{\mathrm{justified}}}{N_{\mathrm{statements}}}.
\]
The rubric is:
\begin{itemize}
    \item $S_r = 10$: $J=1$, all statements justified;
    \item $S_r \in [7,9]$: $J \ge 0.8$, mostly rigorous;
    \item $S_r \in [4,6]$: $0.5 \le J < 0.8$, partial justification;
    \item $S_r \in [0,3]$: $J < 0.5$, predominantly unjustified reasoning.
\end{itemize}

\subparagraph{Logical flow ($S_l$).}
Logical coherence is assessed using the dependency structure of the derivation:
\begin{itemize}
    \item $S_l = 10$: the derivation forms a clear directed acyclic graph (DAG), meaning a non-circular dependency structure in which each step follows from prior steps;
    \item $S_l \in [7,9]$: mostly sequential with minor organizational gaps;
    \item $S_l \in [4,6]$: unclear dependencies or partially disconnected reasoning;
    \item $S_l \in [0,3]$: non-sequential reasoning, circularity, or disconnected argument structure.
\end{itemize}

\subparagraph{Quality of justification ($S_j$).}
Explanation depth is measured by the average reasoning-chain length
\[
R = \mathrm{avg}(\text{explanation steps per result}).
\]
The rubric is:
\begin{itemize}
    \item $S_j = 10$: $R \ge 3$;
    \item $S_j \in [7,9]$: $2 \le R < 3$;
    \item $S_j \in [4,6]$: $1 \le R < 2$;
    \item $S_j \in [0,3]$: $R < 1$.
\end{itemize}

\subparagraph{Completeness ($S_m$).}
Coverage is measured by
\[
M = \frac{N_{\mathrm{addressed\ requirements}}}{N_{\mathrm{total\ requirements}}}.
\]
The rubric is:
\begin{itemize}
    \item $S_m = 10$: $M=1$;
    \item $S_m \in [7,9]$: $M \ge 0.8$;
    \item $S_m \in [4,6]$: $0.5 \le M < 0.8$;
    \item $S_m \in [0,3]$: $M < 0.5$.
\end{itemize}

\subparagraph{Physical consistency ($S_p$).}
Physical validity is assessed via dimensional analysis, unit propagation, and limiting-case behavior:
\begin{itemize}
    \item $S_p = 10$: dimensionally consistent, correct limits, physically plausible;
    \item $S_p \in [7,9]$: generally consistent with minor issues;
    \item $S_p \in [4,6]$: some unit or limiting-behavior errors;
    \item $S_p \in [0,3]$: unphysical or dimensionally inconsistent results.
\end{itemize}

\paragraph{Critic score.}
When Critic feedback is included, it is scored separately:
\[ \begin{aligned} S_{\mathrm{critic}} &= C_a + C_d + C_f + C_t + C_c + 
  C_h + C_v + C_o, \\ S_{\mathrm{critic}} &\in [0,100] \end{aligned} \]  
with:
\begin{align*}
C_a &\in [0,20] && \text{accuracy of identification},\\
C_d &\in [0,15] && \text{depth of analysis},\\
C_f &\in [0,15] && \text{constructive feedback quality},\\
C_t &\in [0,15] && \text{technical understanding},\\
C_c &\in [0,10] && \text{clarity of communication},\\
C_h &\in [0,10] && \text{comprehensiveness},\\
C_v &\in [0,10] && \text{pedagogical value},\\
C_o &\in [0,5]  && \text{objectivity and fairness}.
\end{align*}
This score is recorded but does not feed the pipeline decisions or the
present analysis. In the dynamical language of \cref{app:dynamics}, it
could become a second observable alongside the Actor score, separating
poor feedback from useful feedback that the Actor fails to take up. We
leave this two-observable analysis for follow-up work.

\subparagraph{Critic accuracy ($C_a$).}
Error-detection quality is quantified using precision and recall,
\[
P = \frac{TP}{TP+FP},
\qquad
R = \frac{TP}{TP+FN}.
\]
High scores require both high precision and high recall.

\subparagraph{Critic depth ($C_d$).}
Depth of analysis is measured by
\[
D = \frac{N_{\mathrm{examined\ steps}}}{N_{\mathrm{total\ steps}}}.
\]

\subparagraph{Constructiveness ($C_f$).}
Actionability is measured by
\[
A = \frac{N_{\mathrm{actionable\ suggestions}}}{N_{\mathrm{total\ suggestions}}}.
\]

\subparagraph{Technical understanding ($C_t$).}
Conceptual accuracy is measured by
\[
T = \frac{N_{\mathrm{correct\ concepts}}}{N_{\mathrm{referenced\ concepts}}}.
\]

\subparagraph{Comprehensiveness ($C_h$).}
Coverage of relevant issues is measured by
\[
H = \frac{N_{\mathrm{addressed\ aspects}}}{N_{\mathrm{problem\ aspects}}}.
\]

\paragraph{Pass criterion.}
A single Actor turn is marked as passing only if all of the following hold:
\begin{equation*}
\mathrm{Pass}
=
\Bigl(\mathrm{Eq}(A,R)=1\Bigr)
\wedge
\Bigl(S_{\mathrm{actor}} \ge 80\Bigr)
\wedge
\Bigl(S_c \ge 40\Bigr).
\end{equation*}
In implementation, a non-equivalent final result always forces
$\mathrm{Pass}=0$. A run is then counted as converged if at least one
of its iterations satisfies this turn-level pass rule under the scoring
Judge (\cref{sec:eval}).

\paragraph{Reference-free use.}
For known textbook problems, correctness and convergence rely on the
reference answer through $\mathrm{Eq}(A,R)$. In a future open-problem
setting this part of the rubric would have to be replaced by external
validation: symbolic checks, limiting cases, simulations, experiments,
theorem-prover output, or expert consistency review. The process
criteria --- mathematical rigor, logical flow, justification quality,
completeness relative to the stated task, and physical consistency ---
remain meaningful without a reference solution. By contrast,
``correctness'', Critic error-identification accuracy, and the present
pass rule are reference-dependent and would become measures of
constraint satisfaction rather than answer matching.
In the dynamical-systems language of \cref{app:dynamics}, removing the
reference solution would keep the score projection but alter the drift,
since the Critic's directional pull is currently reference-conditioned.

\paragraph{Error taxonomy.}
The Judge also records categorical errors. At the solution level, major issues are classified into:
\begin{footnotesize}
\begin{flushleft}
\texttt{COMPUTATIONAL\_ERROR}, \texttt{CONCEPTUAL\_ERROR},
\texttt{METHODOLOGICAL\_ERROR}, \texttt{DIMENSIONAL\_ERROR},
\texttt{BOUNDARY\_CONDITION\_ERROR}, \texttt{CONVERGENCE\_ERROR}, and
\texttt{APPROXIMATION\_ERROR}.
\end{flushleft}
\end{footnotesize}
Most categories are self-explanatory. The less obvious labels are:
\texttt{METHODOLOGICAL\_ERROR}, for using an inappropriate setup or
solution method; \texttt{BOUNDARY\_CONDITION\_ERROR}, for missing or
misapplied boundary/normalization conditions; \texttt{CONVERGENCE\_ERROR},
for invalid limiting, integral, or series-convergence reasoning; and
\texttt{APPROXIMATION\_ERROR}, for unjustified expansions or dropped terms.

\paragraph{Binary error flags.}
The Judge records the following binary diagnostic flags:
\begin{footnotesize}
\begin{flushleft}
\texttt{SIGN\_ERROR}, \texttt{MISSING\_TERM},
\texttt{PRODUCT\_RULE\_ERROR}, \texttt{CHAIN\_RULE\_ERROR},
\texttt{BOUNDARY\_CONDITION\_MISAPPLIED}, \texttt{UNIT\_MISMATCH},
\texttt{ALGEBRA\_SIMPLIFY\_FAIL}, \texttt{LIMIT\_ERROR},
\texttt{NORMALIZATION\_ERROR}, \texttt{INCOMPLETE\_EXPR},
\texttt{TEST\_FAIL}, \texttt{TENSOR\_INDEX\_ERROR},
\texttt{DIMENSIONAL\_CONSISTENCY}, \texttt{SYMMETRY\_PRESERVATION},
\texttt{COMBINATORIAL\_FACTORS}, \texttt{REGULARIZATIONAL\_CONSISTENCY},
\texttt{MISSING\_JUSTIFICATION\_NON\_TRIVIAL\_STEPS},
\texttt{LIMITING\_BEHAVIOR\_FAIL},
\texttt{POSITIVITY\_AND\_REALITY\_CONSTRAINTS\_VIOLATION},
\texttt{VIOLATION\_OF\_WARD\_IDENTITIES}.
\end{flushleft}
\end{footnotesize}
Most labels are self-explanatory. The less standard ones correspond to
failed problem-specific checks: \texttt{TEST\_FAIL} marks failure of an
explicit validation test, \texttt{COMBINATORIAL\_FACTORS} covers missing
symmetry or counting factors, and \texttt{VIOLATION\_OF\_WARD\_IDENTITIES}
marks failure of gauge/current-conservation constraints. The names are
legacy JSON keys, so some positive-sounding labels are interpreted as
diagnostic indicators when set to true.

\paragraph{Missing-content labels.}
The Judge may also assign missing-content labels to the Actor solution:
\begin{footnotesize}
\begin{flushleft}
\texttt{UNJUSTIFIED\_CLAIMS}, \texttt{INCORRECT\_RESULT},
\texttt{INCOMPLETE\_DERIVATION}, \texttt{MISSING\_BOUNDARY\_CONDITIONS},
\texttt{DIMENSIONAL\_INCONSISTENCY}, and \texttt{CONVERGENCE\_FAILURE}.
\end{flushleft}
\end{footnotesize}

\paragraph{Progress signal.}
To assess whether Critic intervention changed the solution appreciably, the Judge also records an Actor--Critic progress indicator:
\[
\mathrm{Progress} =
\mathbf{1}\!\left(
\left|S_{\mathrm{current}} - S_{\mathrm{previous}}\right| > \delta
\right),
\]
where $\delta$ is a threshold, typically set to $5$ points.

\paragraph{Implementation note.}
The Judge is required to return a strict JSON object containing the above fields. If the returned structure is invalid or unparsable, the system defaults to a failing verdict and records the malformed response for debugging.

\begin{figure*}[t]
  \centering
  \includegraphics[width=\textwidth]{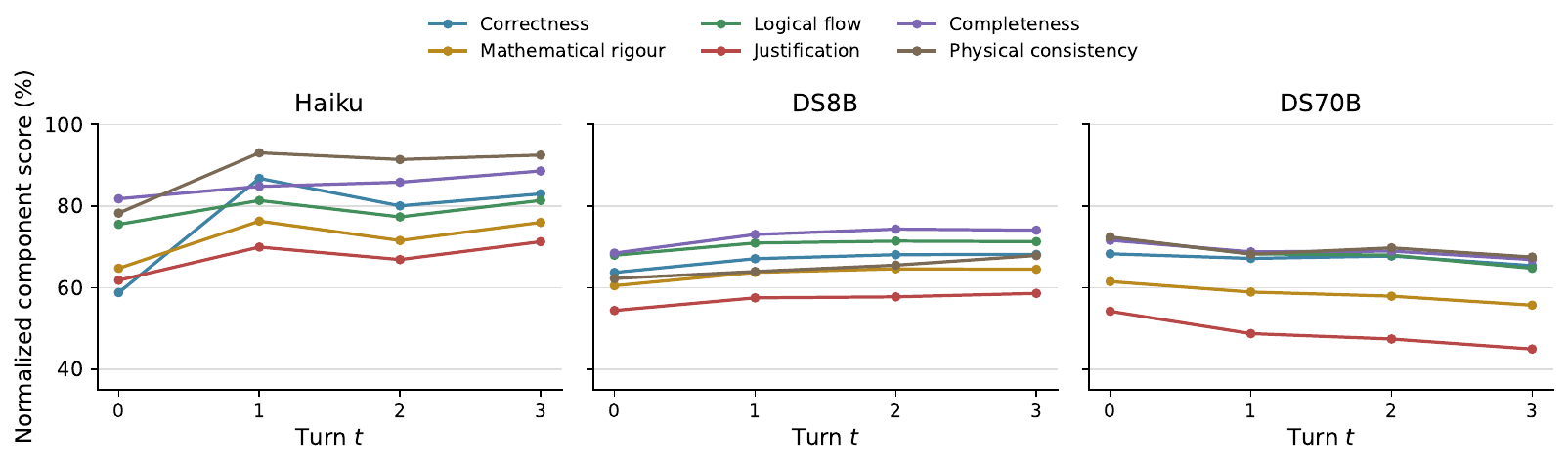}
  \caption{Actor score-component evolution under QWQ scoring. Each
    component is normalized to its rubric maximum before averaging.
    Haiku uses the two Peskin~\& Schroeder problems; DS8B and DS70B use all three
    problems. This diagnostic decomposes the Judge score and is not an
    additional aggregate evaluation metric.}
  \label{fig:app_score_components}
\end{figure*}

The component-score diagnostic in \Cref{fig:app_score_components}
decomposes the Actor score into the six rubric criteria defined in
\cref{sec:eval}. Each line is normalized to the maximum value of that
criterion, so the plot shows which parts of the Judge score move during
the dialogue rather than how many raw rubric points each criterion
contributes. This is not an additional aggregate evaluation metric used
in the statistical tests; it is a decomposition of the score $\bar s$
summarizes. The figure is descriptive: because SCALAR stops runs after
a passing verdict, later turns average over the subset of runs still
active at that turn.

\section{Evaluation Metrics and Statistical Procedures}
\label{app:metrics}

This appendix supplements \cref{sec:eval} with details not in the main
text: the precise pass rule used in the convergence count, edge cases
for runs that terminate early, and the specific statistical test
implementations. The scoring rubric itself (the six Judge dimensions,
pass criterion, and error flags) is described in \cref{app:scoring}.

\paragraph{Iteration range and early stopping.}
The iteration cap is $T_{\max}=4$ for the DeepSeek-family runs and
$T_{\max}=5$ for the Haiku runs; a run terminates earlier at
$T<T_{\max}$ when the Judge used during generation issues a passing
verdict or when the score-stagnation trigger fires. The reported evaluation
metrics use the actual $T$ of
each run, so $\bar s_i$ and $g_i$ are comparable across runs of different
length. Runs that terminate after a single iteration ($T{=}1$) have
$g_i{=}0$ by construction.

\paragraph{Pass rule behind the convergence rate.}
The turn-level pass rule from \cref{app:scoring} requires final-answer
equivalence with the reference, total Actor score~$\ge 80$, and
correctness component~$\ge 40$; a non-equivalent final answer forces
$\mathrm{Pass}{=}0$ regardless of the other components. A run is
counted as converged if any iteration of the dialogue satisfies this
rule under the scoring Judge. For runs whose original loop was driven
by the same Judge that scores them, this is equivalent to a final-turn
pass; for re-scored transcripts the loop length is fixed by the
original Judge and the scoring Judge can mark a non-terminal iteration
as the convergence event.

\paragraph{Statistical tests.}
Each run contributes one independent observation for each response
variable. All tests are two-sided, non-parametric (no Gaussianity
assumption on the score distributions), and use the default
\texttt{scipy.stats} implementations, with the classical tests cited
below for provenance.

Unless explicitly stated otherwise, the statistical response variables are
the raw run-level evaluation quantities $\bar s_i$, $g_i$, and $r_i$.
The problem-normalized contrasts $D_{\bar s}$ and $D_R$ are descriptive
summaries used in \Cref{fig:cross_system}; they are not inputs to the
$p$-values reported in the Results.

Throughout the paper we use the conventional
$p < 0.05$ threshold to flag ``statistically significant'' results:
this means the test statistic is at least as extreme as the $95$th
percentile of its null distribution --- equivalently, under the null
hypothesis (no real effect) a result this surprising or more would
occur less than $1$ in $20$ times;
smaller $p$-values carry stronger evidence against the null. In the main
text, each quoted $p$-value belongs to the statistical test named in the
same sentence, usually a Kruskal--Wallis omnibus test across Critic
strategies or Actor personas. When a re-scoring check is mentioned
without a new $p$-value, it is used only as a sensitivity check on the
direction of the conclusion, not as a separate headline test.

\begin{itemize}
\item \textbf{Wilcoxon signed-rank} on paired $(s_0, s_{T-1})$ scores
  (\texttt{scipy.stats.wilcoxon}; \citealp{wilcoxon1945individual}).
  A paired-sample test that asks
  whether the median of the within-run gains
  $g_i=s_{i,T_i-1}-s_{i,0}$ differs from zero; we use it to assess whether the multi-turn dialogue
  induces a non-zero gain over the population of runs.
\item \textbf{Kruskal--Wallis $H$-test} on $\bar s$ across the five
  Critic strategies (\texttt{scipy.stats.kruskal};
  \citealp{kruskal1952use}). A rank-based
  multi-group test that asks whether samples from $\ge 2$ groups
  come from a common distribution; we use it to assess whether the
  Critic feedback strategy label has any overall effect on $\bar s$, applied
  separately for each Actor--Critic pairing.
\item \textbf{Mann--Whitney $U$-test} for specific pairwise Critic feedback strategy
  comparisons (\texttt{scipy.stats.mannwhitneyu};
  \citealp{mann1947test}). A two-sample
  rank-based test that asks whether values from one group tend to be
  larger than values from another; we use it to assess whether a
  particular pair of Critic feedback strategies (\emph{e.g.},\ pedagogical vs.\ strict)
  differ. We report uncorrected $p$-values; where multiple comparisons
  contribute to the same claim we flag this alongside the relevant
 result. We do not use rank correlations between five Critic feedback strategy means as
 evidential tests, because with only five ranks they are too coarse to
 support an independent statistical claim.

\end{itemize}

\paragraph{Re-scoring protocol.}
Re-scoring Judges score the stored Actor--Critic transcripts using the same
Judge prompt (\cref{app:prompts}) but a different Judge LLM; the Actor
and Critic are not re-run, and the original stopping decisions are
preserved. The Haiku transcripts are re-scored by QWQ and by
DS70B in addition to the primary Sonnet scoring, and the DS70B
and DS8B transcripts are re-scored by DS70B in
addition to the primary QWQ scoring. This supports the cross-Judge
consistency checks reported in \cref{sec:results}.

\section{Score-Update Field and Markov Projection}
\label{app:dynamics}

\begin{figure*}[t]
  \centering
  \includegraphics[width=\textwidth]{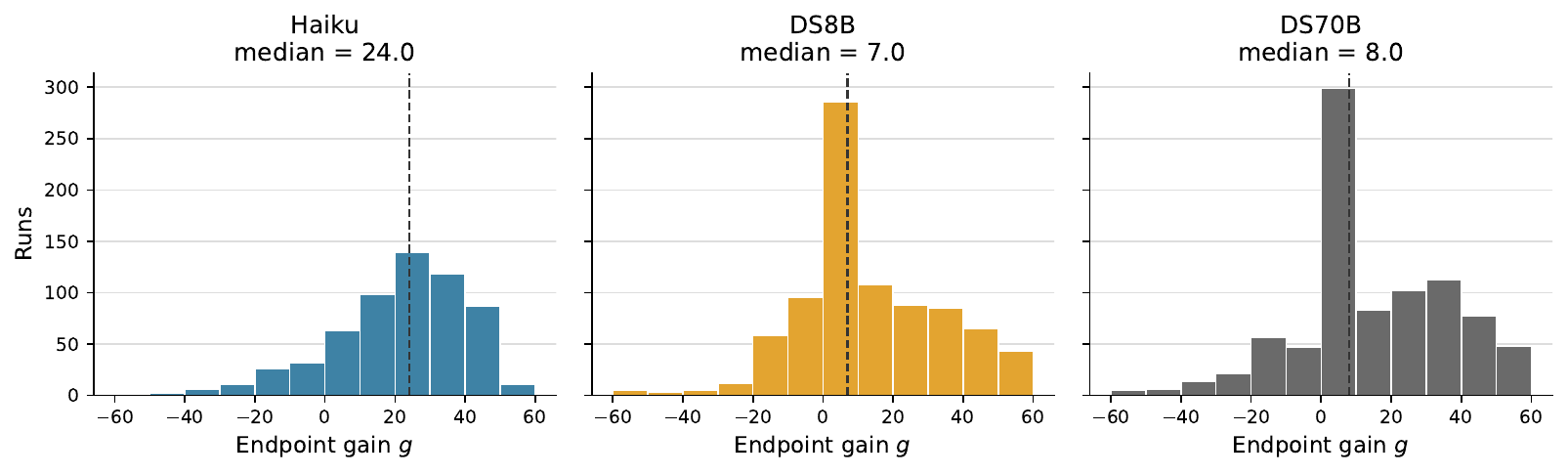}
  \caption{Endpoint gain distributions under QWQ scoring. Haiku uses the
    two Peskin~\& Schroeder problems \textcolor{blue}{($n{=}600$)}, while DS8B and DS70B use all three
    problems ($n{=}900$ each). Dashed vertical lines mark the median
    gain.}
  \label{fig:app_gain_dist}
\end{figure*}

The score-update curves in \cref{fig:ds_param_scaling} estimate a
one-step score drift; they are not inputs to the hypothesis tests in the
Results. Let $X_t$ denote the full SCALAR state after Actor turn~$t$:
problem, persona, Critic feedback strategy, transcript, scoring state, and
model-sampling rule. Conditional on the experimental cell, the algorithm
induces a one-step transition
\[
p(X_{t+1}\mid X_t,X_{t-1},\ldots)=p(X_{t+1}\mid X_t).
\]
The scalar Judge score is a projection $s_t=S(X_t)$, and the useful
question is how much of the dialogue dynamics remains visible in this
one-dimensional coordinate. For run $i$, let $s_{i,t}$ be the Judge
score after Actor turn~$t$ and define
$\Delta s_{i,t}=s_{i,t+1}-s_{i,t}$. For a score bin $B_b$, we estimate
the empirical drift
\[
\hat v_b =
\frac{1}{|\mathcal{T}_b|}
\sum_{(i,t)\in\mathcal{T}_b} \Delta s_{i,t},
\qquad
\mathcal{T}_b=\{(i,t):s_{i,t}\in B_b\}.
\]
The shaded bands in \cref{fig:ds_param_scaling} are bootstrap confidence
intervals for $\hat v_b$ over runs. When the fitted curve crosses zero,
we denote the crossing by $s^\ast$ and call it a projected fixed point. This
means only that the observed next-turn update vanishes on average near
that score; it is not necessarily a complete fixed point of the full
recorded dialogue state.

One can also estimate the local fluctuation scale
\[
\hat D_b=\frac{1}{2}\operatorname{Var}_{(i,t)\in\mathcal{T}_b}
\!\left(\Delta s_{i,t}\right),
\]
the empirical analogue of diffusion for the projected score. Here
stochasticity comes both from nonzero-temperature LLM sampling and from
projecting many transcript states to the same score. We do not fit a
continuous Fokker--Planck or Smoluchowski model here: the trajectories
are short, early stopping censors the active ensemble, and passing
depends on final-answer equivalence rather than on a score threshold
alone. A future temperature sweep could separate drift-dominated
improvement from noise-assisted escape out of low-drift regions.

Discarding the transcript gives a score-only Markov closure,
\[
p(s_{t+1}\mid s_t,s_{t-1},\ldots)\approx p(s_{t+1}\mid s_t),
\]
after conditioning on the experimental cell. This does not follow from
the transcript-level Markov property: two transcripts with the same score
can contain different physics errors. As an exploratory DS70B/QWQ check,
we regressed the residual $\Delta s_t-\hat v(s_t)$ on the previous score
$s_{t-1}$ within each problem; the fitted memory terms were small compared
with their standard errors. Thus the present data are
consistent with the score-only closure, while longer held-out
trajectories are needed for predictive validation.

The predictive version would fit a discrete transition kernel over score
bins with an absorbing pass state
$\mathcal{P}=\{s\geq 80\ \mathrm{and\ final\ answer\ equivalent}\}$.
On held-out trajectories this kernel could predict score histograms,
convergence rates, or mean remaining turns; in the absorbing-chain
idealization, the non-pass bins obey $\tau=(I-K_{BB})^{-1}\mathbf{1}$.

\end{document}